\documentclass{article} 
\usepackage{times}

\usepackage{amsmath,amsfonts,bm}

\def\eqref#1{equation~\ref{#1}}

\def\1{\bm{1}}

\DeclareMathAlphabet{\mathsfit}{\encodingdefault}{\sfdefault}{m}{sl}
\SetMathAlphabet{\mathsfit}{bold}{\encodingdefault}{\sfdefault}{bx}{n}

\usepackage{hyperref}
\usepackage{url}
\usepackage{breakurl}

\usepackage{wrapfig}
\usepackage{graphicx}
\usepackage{algorithmic}
\usepackage{algorithm}
\usepackage{booktabs} 
\usepackage{caption}
\usepackage{subcaption}
\usepackage{amsmath,amssymb,amsthm}
\usepackage{nicefrac}
\usepackage{xcolor}
\usepackage{enumitem}

\usepackage{colortbl}
\usepackage{soul}
\usepackage[framemethod=tikz]{mdframed}

\usepackage{enumitem}
\usepackage{tabularx}
\newtheorem{theorem}{Theorem}[section]
\newtheorem{proposition}[theorem]{Proposition}
\newtheorem{remark}[theorem]{Remark}

\usepackage[accepted]{icml2024}

\begin{document}

\twocolumn[
\icmltitle{Disentanglement Learning via Topology}

\icmlsetsymbol{equal}{*}

\begin{icmlauthorlist}
\icmlauthor{Nikita Balabin}{equal,skoltech}
\icmlauthor{Daria Voronkova}{equal,skoltech,airi}
\icmlauthor{Ilya Trofimov}{skoltech}
\icmlauthor{Evgeny Burnaev}{skoltech,airi}
\icmlauthor{Serguei Barannikov}{skoltech,cnrs}
\end{icmlauthorlist}

\icmlaffiliation{skoltech}{Skolkovo Institute of Science and Technology, Moscow, Russia}
\icmlaffiliation{airi}{AIRI, Moscow, Russia}
\icmlaffiliation{cnrs}{CNRS, IMJ, Paris Cité University}

\icmlcorrespondingauthor{Nikita Balabin}{nikita.balabin@skoltech.ru}
\icmlcorrespondingauthor{Daria Voronkova}{Darya.Voronkova@skoltech.ru}

\icmlkeywords{Machine Learning, ICML}

\vskip 0.3in
]

\printAffiliationsAndNotice{\icmlEqualContribution}

\begin{abstract}

We propose TopDis (Topological Disentanglement), a method for learning disentangled representations via adding a multi-scale topological loss term. Disentanglement is a crucial property of data representations substantial for the explainability and robustness of deep learning models and a step towards high-level cognition. The state-of-the-art methods are based on VAE and encourage the joint distribution of latent variables to be factorized. We take a different perspective on disentanglement by analyzing topological properties of data manifolds. In particular, we optimize the topological similarity for data manifolds traversals. To the best of our knowledge, our paper is the first one to propose a differentiable topological loss for disentanglement learning. Our experiments have shown that the proposed TopDis loss improves disentanglement scores such as MIG, FactorVAE score, SAP score, and DCI disentanglement score with respect to state-of-the-art results while preserving the reconstruction quality. Our method works in an unsupervised manner, permitting us to apply it to problems without labeled factors of variation. The TopDis loss works even when factors of variation are correlated. Additionally, we show how to use the proposed topological loss to find disentangled directions in a trained GAN.

\end{abstract}

\section{Introduction}
Learning disentangled representations is a fundamental challenge in deep learning, as it has been widely recognized that achieving interpretable and robust representations is crucial for the success of machine learning models \citep{bengio2013representation}.
Disentangled representations, in which each component of the representation corresponds to one factor of variation \citep{desjardins2012disentangling, bengio2013representation, cohen2014learning, kulkarni2015deep, chen2016infogan, higgins2017beta, tran2021group, feng2020dual, gonzalez2018image}, have been shown to be beneficial in a variety of areas within machine learning.
One key benefit of disentangled representations is that they enable effective domain adaptation, which refers to the ability of a model to generalize to new domains or tasks. 
Studies have shown that disentangled representations can improve performance in unsupervised domain adaptation \citep{yang2019unsupervised, peebles2020hessian, zou2020joint}. 
Additionally, disentangled representations have been shown to be useful for zero-shot and few-shot learning, which are techniques for training models with limited labeled data \citep{bengio2013representation}.
Disentangled representations have also been shown to enable controllable image editing, which is the ability to manipulate specific aspects of an image while keeping the rest of the image unchanged \citep{wei2021orthogonal,wang2021geometry}. This type of control can be useful in a variety of applications, such as image synthesis, style transfer, and image manipulation.

\begin{figure}
\centering
\includegraphics[width=0.49\textwidth]{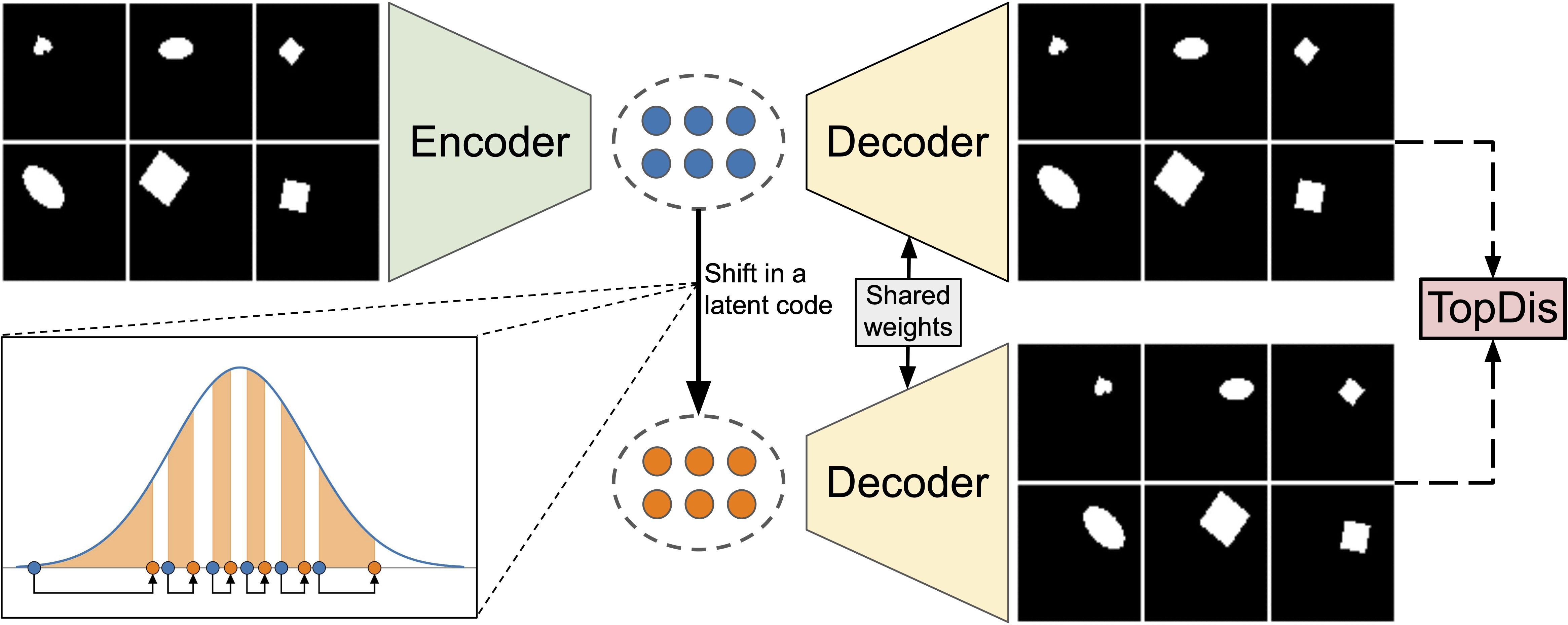}

\caption{The TopDis pipeline process involves the following steps: encoding a batch of data samples, applying shift in a latent code, decoding both the original and the shifted latents, and finally calculating the TopDis loss between the two resulting point clouds, for details see Section \ref{sec:method}.}
\label{fig:td_pipeline}
\vskip-.2in
\end{figure}

Furthermore, disentangled representations are also believed to be a vital component for achieving high-level cognition. High-level cognition refers to the ability of a model to understand and reason about the world, and disentangled representations can play a key role in achieving this goal \citep{bengio2018cognition}.

One line of research for finding disentangled representations is to modify the Variational Autoencoder (VAE) \citep{kingma2013auto} using some intuition, formalizing statistical independence of latent components \citep{higgins2017beta, chen2018isolating, kim2018disentangling}, or the group theory based definition of disentanglement \citep{yang2021towards}.
Another line is to modify Generative Adversarial Networks (GANs) \citep{goodfellow2014generative, chen2016infogan, lin2020infogan, peebles2020hessian, wei2021orthogonal} to enforce the change in a particular component being predictable or independent in some sense from other components.

At the same time, \citet{locatello2019challenging} stated the impossibility of fully unsupervised learning of disentangled representation with a statistical approach. However, empirical evidence shows that disentanglement learning is possible, probably due to inductive bias either in the model or the dataset \citep{michlooverlooked, rolinek2019variational}. We follow \citet{higgins2018towards}, Section 3, where it is pointed out that one can achieve disentanglement w.r.t. the natural decomposition through active intervention, which in our case takes the form of the proposed group(oid) action shifts. Also, our work is based on exploring topological properties of a data manifold. Thus, statistical arguments of \citet{locatello2019challenging} do not apply in our case.

The proposed approach is grounded in the manifold hypothesis \citep{goodfellow2016deep} which posits that data points are concentrated in a vicinity of a low-dimensional manifold. For disentangled representations, it is crucial that the manifold has a specific property, namely, small topological dissimilarity between a point cloud given by a batch of data points and another point cloud obtained via the symmetry group(oid) action shift along a latent space axis.
To estimate this topological dissimilarity, we utilize the tools from topological data analysis \citep{Barannikov1994, chazal2017introduction}.
We then develop a technique for incorporating the gradient of this topological dissimilarity measure into the training of VAE-type models.

Our contributions are the following:
\begin{itemize}[topsep=0pt,noitemsep,nolistsep, partopsep=0pt,parsep=0ex,leftmargin=2ex]
\item We propose TopDis (Topological Disentanglement), a method for unsupervised learning of disentangled representations via adding to a VAE-type loss the topological objective;
\item Our approach uses group(oid) action shifts preserving the Gaussian distribution;
\item We improve the reconstruction quality by applying gradient orthogonalization;
\item Experiments show that the proposed TopDis loss improves disentanglement metrics (MIG, FactorVAE score, SAP score, DCI disentanglement score) with respect to state-of-the-art results. Our method works even when factors of variation are correlated.
\end{itemize}

We release out code: \url{https://github.com/nikitabalabin/TopDis}

\section{Related Work} 

In generative models, disentangled latent space can be obtained by designing specific architectures of neural networks \citep{karras2019style} or optimizing additional loss functions. While the latter approach can admit supervised learning \citep{kulkarni2015deep, kingma2014semi, paige2017learning, mathieu2016disentangling, denton2017unsupervised}, the most challenging but practical approach is unsupervised learning of disentangled representations since the underlying factors of variation are typically unknown for real data.

The Variational Autoencoder (VAE) \citep{kingma2013auto}, a widely used generative model, is not able to achieve disentanglement alone. To address this limitation, researchers have proposed different modifications of VAE such as $\beta$-VAE \citep{higgins2017beta}, which aims to promote disentanglement by increasing the weight of Kullback–Leibler (KL) divergence between the variational posterior and the prior. To overcome the known trade-off between reconstruction quality and disentanglement \citep{sikka2019closer}, some researchers have proposed to use the concept of total correlation. In $\beta$-TCVAE \citep{chen2018isolating}, the KL divergence between the variational posterior and the prior is decomposed into three terms: index-code mutual information, total correlation (TC), and dimension-wise KL. The authors propose to penalize the TC as the most important term for learning disentangled representations. However, estimation of the three terms of decomposition is challenging, and the authors propose a novel framework for training with the TC-decomposition using minibatches of data. The authors of FactorVAE \citep{kim2018disentangling} proposed an additional discriminator which encourages the distribution of latent factors to be factorized and hence independent across the dimensions without significantly reducing the reconstruction loss. Recently \citet{estermann2023dava} proposed DAVA, an adversarial framework for learning disentangled representations with dynamical hyperparameters tuning. \citet{moor2020topological} proposed a topological loss term for autoencoders that helps harmonise the topology of the data
space with the topology of the latent space.

\citet{locatello2019challenging} conduct a comprehensive empirical evaluation of a large amount of existing models for learning disentangled representations, taking into account the influence of hyperparameters and initializations.
They find that the FactorVAE method achieves the best quality in terms of disentanglement and stability while preserving the reconstruction quality of the generated images.

\begin{samepage}
Approaches to interpretation of neural embeddings are developed in \citep{bertolini2022explaining, zhang2018interpretable, zhou2018interpreting}. In the work \citep{shukla2018geometry}, the authors study a geometry of deep generative models for disentangled representations. Tools of topological data analysis were previously applied to disentanglement evaluation \citep{barannikov2021representation, zhou2020evaluating}.
In \citep{barannikov2021representation}, topological dissimilarity in data submanifolds corresponding to slices in latent space for a simple synthetic dataset was compared.
 \end{samepage}

\section{Background}
\subsection{Variational Autoencoder}

The Variational Autoencoder (VAE) \citep{kingma2013auto} is a generative model that encodes an object $x_n$ into a set of parameters of the posterior distribution $q_\phi(z|x_n)$, represented by an encoder with parameters $\phi$. Then it samples a latent representation from this distribution and decodes it into the distribution $p_\theta(x_n|z)$, represented by a decoder with parameters $\theta$. The prior distribution for the latent variables is denoted as $p(z)$. In this work, we consider the factorized Gaussian prior $p(z) = N(0,I)$, and the variational posterior for an observation is also assumed to be a factorized Gaussian distribution with the mean and variance produced by the encoder. The standard VAE model is trained by minimizing the negative Evidence Lower Bound (ELBO) averaged over the empirical distribution:
\begin{equation*}
\mathcal{L}_{VAE}=\mathcal{L}_{rec}+\mathcal{L}_{KL}, \text{ where}
\end{equation*}
\begin{equation*}
\mathcal{L}_{rec}=-\frac{1}{N} \sum_{n=1}^{N}\mathbb{E}_{q}\left[\log p_\theta\left(x_{n} \mid  z\right)\right],
\end{equation*}
\begin{equation*}
\mathcal{L}_{KL}=-\frac{1}{N} \sum_{n=1}^{N}\operatorname{KL}\left(q_\phi\left(z \mid  x_{n}\right)|| \, p(z)\right).
\end{equation*}

\begin{figure}
    \centering
    \includegraphics[width=0.35\textwidth]{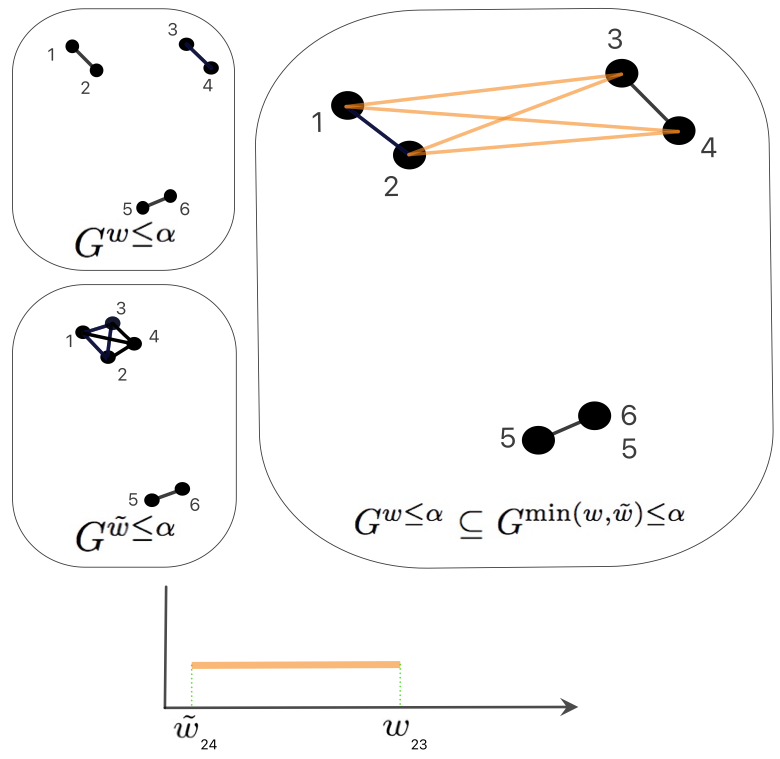}
    \caption{An example of RTD calculation.}
    \label{fig:3to2clusters}
\vskip-0.4cm
\end{figure}

\subsection{Representation Topology Divergence}

\label{sec:method}
\begin{figure*}[t!]
\centering
\begin{subfigure}[t]{0.49\textwidth}
    \centering
    \includegraphics[width=0.75\textwidth]{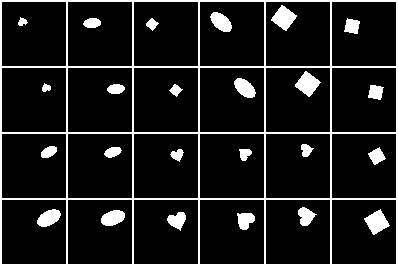}
    \caption{Example of traversals in dSprites dataset.}
    \label{fig:rtd_traverse}
\end{subfigure}
\begin{subfigure}[t]{0.49\textwidth}
    \centering
    \includegraphics[width=0.60\textwidth]{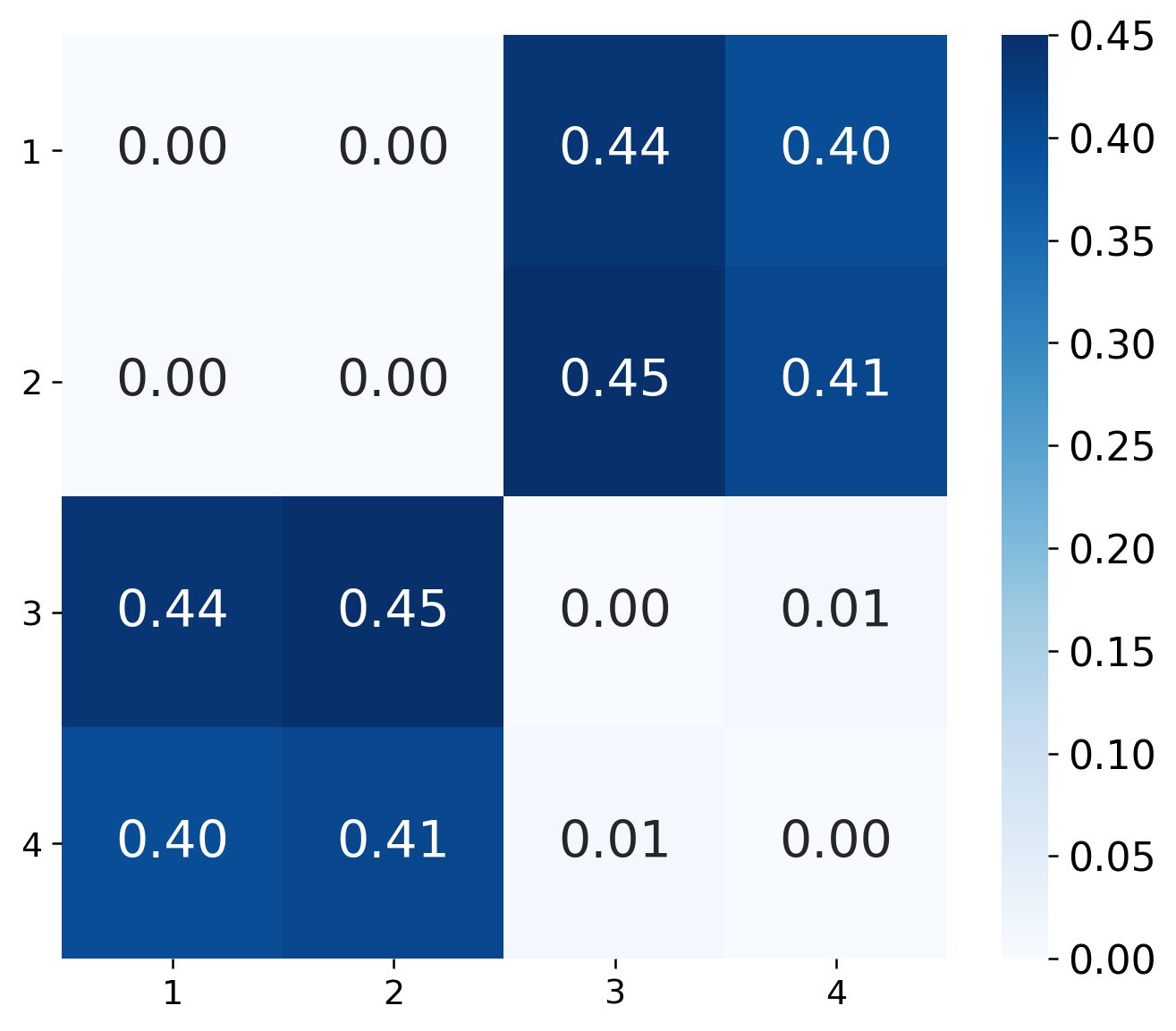}
    \caption{RTD between point clouds represented as rows in Figure \ref{fig:rtd_traverse}.}
    \label{fig:rtd_rows}
\end{subfigure}%
\caption{Left: rows represent point clouds (mini-batches). The 1st row represents a random batch of samples; the 2nd row is obtained by equally shifting samples from the 1st row to the right; the 3rd row is placed the same as 2nd, but all objects are randomly transformed; the 4th row is a scaling of samples from 3rd row. The RTD value between the 1st and 2nd point clouds is zero, as RTD between the 3rd and 4th rows. While RTD between the 2nd and 3rd rows is large because the topological structures of these two clouds are not similar.}
\vskip-.2in
\end{figure*}

Representation Topology Divergence (RTD) \citep{barannikov2021representation} is a topological tool comparing two point clouds $X, \tilde{X}$ with one-to-one correspondence between points. RTD compares multi-scale topological features together with their localization.  The distances inside clouds $X, \tilde{X}$ define  two weighted graphs ${G}^{{w}}$, ${G}^{\tilde{w}}$ with the same vertex set $X$, $w_{AB}={\operatorname{dist}}(A,B)$, $\tilde{w}_{AB}={\operatorname{dist}} (\tilde{A},\tilde{B})$.
For a threshold $\alpha$,  the graphs $G^{w\leq \alpha}$, $G^{\tilde{w}\leq\alpha}$ are the $\alpha$-neighborhood graphs of $X$ and  
 $\tilde{X}$. 
RTD tracks the differences in multi-scale topology between $G^{w\leq \alpha}$, $G^{\tilde{w}\leq\alpha}$ by comparing them with the graph $G^{\min(w,\tilde{w})\leq\alpha}$, which  contains an edge between vertices 
$A$ and $B$ iff an edge between $A$ and $B$ is present in either $G^{w\leq \alpha}$ or $G^{\tilde{w}\leq\alpha}$.
Increasing $\alpha$ from $0$ to the diameter of $X$, the connected components in $G^{w\leq\alpha}(X)$ change from $|X|$ separate vertices to one connected component with all vertices.
Let $\alpha_1$ be the scale at which a pair of connected components $C_1, C_2$ of $G^{w\leq \alpha}$ becomes joined into one component in $G^{\min(w,\tilde{w})\leq\alpha}$.
Let at some $\alpha_2 > \alpha_1$, the components $C_1, C_2$ become also connected  in $G^{w\leq \alpha}$. 
\textit{R-Cross-Barcode$_1(X,\tilde{X})$} is the multiset of intervals like $[\alpha_1, \alpha_2]$, see Figure \ref{fig:3to2clusters}. Longer intervals indicate in general the essential topological discrepancies between $X$ and $\tilde{X}$.
By definition, RTD is the half-sum of interval lengths in R-Cross-Barcode$_1(\tilde{X}, X)$ and R-Cross-Barcode$_1(X, \tilde{X})$. By simplicial complexes based formal definition, R-Cross-Barcode  is the barcode of the graph $\hat{\mathcal{G}}^{w, \tilde{w}}$ from \citep{barannikov2021representation}, see  Appendix \ref{app:formal_rtd}.

Figure \ref{fig:3to2clusters} illustrates the calculation of RTD. The case  with  
three clusters in $X$ merging into two clusters in $\tilde{X}$ is shown.
Edges of $G^{\tilde{w}\leq\alpha}$ not in $G^{w\leq\alpha}$, 
are colored in orange. In this example there are exactly four edges of different weights  $(13),(14),(23),(24)$ in the point clouds $X$ and  $\tilde{X}$.  The unique topological feature in \emph{R-Cross-Barcode$_1(X,\tilde{X})$} in this case  is born at the threshold ${\tilde{w}}_{24}$ when the difference in the cluster structures of the two graphs arises, as the points $2$ and $4$ are in the same cluster at this threshold in $G^{\min(w,\tilde{w})\leq\alpha}$ and not in $G^{w\leq\alpha}$. This feature dies at the threshold $\alpha_2=w_{23}$ since the clusters  $(1,2)$ and $(3,4)$ are merged at this threshold in $G^{w\leq\alpha}$. 

The differentiation of RTD is described in \citep{trofimov2023learning}, see also Appendix \ref{app:rtd_diff}.

\section{Method}

\subsection{Topology-aware Loss for Group(oid) Action}\label{TopGroup}

First, we provide a simple example demonstrating the relevance of topology-aware loss. The disentanglement is illustrated commonly by traversals along axes in latent space. Figure \ref{fig:rtd_traverse} presents an example of various shifts in the latent space for the dSprites dataset with known factors of variations. In Figure \ref{fig:rtd_rows} we demonstrate that transformations in disentangled directions have minimal topological dissimilarities as measured by RTD between two sets of samples. As illustrated by this example, minimization of RTD should favor the decomposition of latent space to disentangled directions.

As we explain below, the minimization of topology divergence is implied by the continuity of the symmetry Lie group(oid) action on data distribution.

\textbf{Definition of VAE-based disentangled  representation}. We propose that the disentangled VAE for data distribution in a space $Y$ consists of (cf. \cite{higgins2018towards}):
\begin{enumerate}[topsep=0pt,noitemsep, partopsep=0pt, parsep=0ex, leftmargin=*]
\item The encoder  $h:Y\to Z$  and the decoder $f:Z\to Y$ neural networks, $Z=\mathbb{R}^n$, maximizing ELBO, with the standard ${N}(0,I)$ prior distribution on $Z$.
\item Symmetry Lie group(oid) actions on distributions in $Y$ and $Z$,  such that the decoder and the encoder are equivariant with respect to group(oid) action, $f(g(z))=g(f(z))$, $h(g(x))=g(h(x))$, $g\in G$.
\item A decomposition  $G=G_1\times\ldots \times G_n$, where $G_i\simeq\mathbb{R}$ are 1-parameter Lie subgroup(oid)s.  We distinguish two cases arising in examples: a) $G_i,G_j$ are commuting with each other: $g_ig_j=g_jg_i$ for $g_i\in G_i, g_j\in G_j$, b) $g_ig_j\approx g_jg_i$ up to higher order $\mathcal{O}(\log||g_i||\log||g_j||)$. 
\item The Lie group(oid) $G$ action on the latent space decomposes and  each $G_i$ acts only on a single latent variable $z_i$, preserving the prior  ${N}(0,1)$ distribution on $z_i$;   it follows from Proposition \ref{prop:groupact} that $G_i$ acts on $z_i$  via the shifts (\ref{shifts}). 
\end{enumerate}

The concept of Lie groupoid is a formalization of continuous symmetries, when, contrary to group in 
\cite{higgins2018towards},
symmetry action is not necessarily applicable to all points. We gather necessary definitions in Appendix \ref{groupoid}.

To relate topological features to disentangled representations, the keys are the \textbf{continuity} and the \textbf{existence of inverse} properties for the Lie action on the support of the model distribution $p_\theta(x)$ in the disentangled model. Transformations with these properties are called homeomorphisms in topology. For a map to be a homeomorphism, a fundamental requirement is that it must induce an isomorphism in homology, meaning it must preserve the topological features. By design, minimizing TopDis guarantees the maximal preservation of the topological features by the Lie action transformations. We prove a  proposition which strengthens this relation further in Appendix \ref{app:motivation}.

In other words, since the Lie group(oid) symmetry action by $g\in G_i$ on the support of data distribution in $Y$ is continuous and invertible, the collection of topological features of its any subset should be preserved under $g$. This can be verified by RTD measuring the discrepancy in topological features at multiple scales of two data samples. Given a sample of images, TopDis loss is given by RTD between the reconstructed sample and the sample reconstructed from the group(oid) shifted latent codes, described in Section \ref{section:shift}. If the two collections of topological features at multiple scales are preserved by $g$, then TopDis loss between the two samples is small.

\vskip-1.1in

\subsection{Group(oid) Action Shifts Preserving the Gaussian Distribution.}
\label{section:shift}
Given a batch of data samples, $X={x_1,\dots,x_N}$, we sample the corresponding latent representations, 
\begin{figure}
\centering
\includegraphics[width=0.4\textwidth]{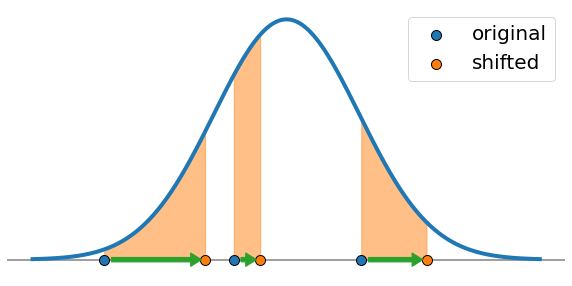}
\vspace{-10pt}
\caption{Shift of real line preserving ${N}(0, 1)$, $C=1/8$. The three orange curvilinear rectangles have the same area: $F(z_{\operatorname{shifted}})-F(z)=1/8$}
\vskip-.1in
\label{fig:probshift}
\end{figure}
$z_n\sim q_{\phi}(z|x_n)$, and the reconstructed samples, $\hat x_n \sim p_{\theta}(x | z_n)$.
To ensure that the shifts in a latent code preserve the prior Gaussian distribution, we propose the shifts defined by the equation:
\begin{equation}\label{shifts}
   z_{\operatorname{shifted}} = F^{-1}(F(z \mid \rho,\sigma^2) + C \mid \rho,\sigma^2) 
\end{equation} 
Shifts in the latent space are performed using the \emph{cumulative} function $F(z\mid \rho,\sigma^2)$ of the Gaussian distribution. The mean value $\rho$ and variance $\sigma^2$ of the distribution are calculated as the empirical mean and variance of the latent code for the given sample of the data, see Algorithm \ref{alg:probshift}. 
\begin{proposition}\label{prop:groupact}
a) For any fixed $\rho,\sigma$, the equation (\ref{shifts}) defines a local action of the additive group $\{C\;\vert\; C\in\mathbb{R}\}$ on real line. b) This abelian group(oid) action preserves the $N(\rho,\sigma^2)$ Gaussian distribution density. c) Conversely, if a local (group(oid)) action of this abelian group preserves the $N(\rho,\sigma^2)$  distribution  then the action is given by formula (\ref{shifts}).
\end{proposition}

See Appendix \ref{proof:groupact} for the proof and more details. 
This shifting is illustrated in Figure \ref{fig:probshift}. 

In the following Proposition \ref{prop:qzi}, we bridge the gap between the proposed definition of disentanglement and the definition originating from \citet{bengio2013representation}. We demonstrate that the group(oid) action shift preserves the aggregate posterior latent code independence as implied by the definition from \citet{bengio2013representation}. This suggests that the two frameworks can be combined in practice.

Let $q(z)$ be the aggregate posterior distribution over the latent space, aggregated over the whole dataset $X$. And let $q(z_i)$ be the similar aggregate distribution over the latent code $z_i$. The formula (\ref{shifts}) is valid and defines symmetry shifts if we replace the standard normal distribution with any distribution over the real line, we use it with the distribution $q(z_i)$ over the $i-$th latent codes.  

\begin{proposition}\label{prop:qzi}
a) If the distribution $q(z)$ is factorized into product $q(z)=\prod_i q(z_i)$, then the shift defined by the formula (1) acting on a single latent code $z_i$ and leaving other latent codes fixed,  preserves the latent space distribution $q(z)$. This defines the $G_i$ groupoid action on $z$ for any $i$, whose action can be then extended to points of the initial dataset $X$ with the help of the decoder-encoder.  b) Conversely, if $q(z)$ is preserved for any $i$ by the shifts acting on $z_i$ and defined via formula (\ref{shifts}) from the distribution $q(z_i),$ then $q(z)=\prod_i q(z_i)$.    
\end{proposition} 
The proof is given in Appendix \ref{proof:qzi}.

\begin{algorithm}[tb]
  \caption{Latent traversal with a shift in the latent space.}
   \label{alg:probshift}
\begin{algorithmic}
   \STATE \hskip-1em {\bfseries Input:} $z \in\mathbb{R}^{N\times d}$ -- an array of latent representations from encoder. $C$ -- the shift value. $F(z \mid \rho,\sigma^2)$ -- the cumulative function for $\mathcal N(\rho,\sigma^2)$ distribution.
    \STATE $i\sim \{1,\dots,d\}$, random choice of latent code
    \STATE $s\sim \{-C, C\}$, random choice of shift direction.
    \STATE $\rho \gets \operatorname{mean}(z^{(i)})$, empirical mean value for $i$-th latent representation along batch.
    \STATE $\sigma^2 \gets \operatorname{var}(z^{(i)})$, empirical variance for the $i$-th latent representation along batch.
    \STATE $p \gets F(z^{(i)} \mid \rho,\sigma^2)$, p-values of batch along $i$-th latent code, $p\in \mathbb{R}^{N}$
    \STATE $\mathcal{J} = \{j \mid p_j + s \in (0, 1)\}$, valid set of the batch elements that can be shifted
    \STATE $z_{\operatorname{original}} \gets \{z_{j} \mid j \in \mathcal{J}\}$, batch of valid original latents
    \STATE $z^{(i')}_{\operatorname{shifted}} \gets z^{(i')}_{\operatorname{original}}$, copy of latents $z^{(i')}_{\operatorname{original}}$, $i'\neq i$ 
    \STATE $z^{(i)}_{\operatorname{shifted}} \gets \{F^{-1}(p_j + s \mid \rho,\sigma^2) \mid j \in \mathcal{J}\}$, apply the shift only along the $i$-th latent code.
   \STATE \hskip-1em {\bfseries Return:} $z_{\operatorname{original}},\;z_{\operatorname{shifted}}$ -- valid original and shifted latents. $z_{\operatorname{original}},\;z_{\operatorname{shifted}} \in\mathbb{R}^{|\mathcal{J}|\times d}$
\end{algorithmic}
\end{algorithm}\vspace{-0.2cm}

\begin{algorithm}[tb]
  \caption{The TopDis loss.}
   \label{alg:tdloss}
\begin{algorithmic}
   \STATE \hskip-1em {\bfseries Input:} $X\in {\mathbb{R}}^{N\times C\times H \times W}$, VAE  parameters $\phi, \theta$, $p\in\{1,2\}$ -- an exponent, $C$ -- the shift scale.
     \STATE $\mu_z, \sigma^2_z \gets q_{\phi}(z|X)$, posterior parameters from encoder given batch $X$.
     
     \STATE $z_{\operatorname{original}},\;z_{\operatorname{shifted}}$ -- valid original and shifted latents, obtained by Algorithm \ref{alg:probshift}
     
     \STATE $\hat{X}_{\operatorname{original}}\sim p_{\theta}(x|z_{\operatorname{original}})$, a reconstruction of initial batch $X$
     \STATE $\hat{X}_{\operatorname{shifted}}\sim p_{\theta}(x|z_{\operatorname{shifted}})$, a generation of modified $X$ after applying shift along some fixed latent code.

     \STATE $\mathcal{L_{TD}} \gets \operatorname{RTD}^{(p)}(\hat{X}_{\operatorname{original}}, \hat{X}_{\operatorname{shifted}})$
   \STATE \hskip-1em {\bfseries Return:} $\mathcal{L_{TD}}$ -- topological loss term.
\end{algorithmic}
\vspace{-0.13cm}
\end{algorithm}


\subsection{The TopDis Loss}

The TopDis loss is calculated using the RTD measure, which quantifies the dissimilarity between two point clouds with one-to-one correspondence. The reconstructed batch of images, $\hat X$, is considered as a point cloud in the $\mathbb{R}^{H\times W\times C}$ space\footnote{For complex images, RTD and the TopDis loss can be calculated in a representation space instead of the pixel space $X$.}; $H$, $W$, and $C$ are the height, width, and number of channels of the images respectively. 
The one-to-one correspondence between the original and shifted samples is realized naturally by the shift in the latent space.
Finally, having the original and shifted point clouds:
\begin{equation}
\hat{X}_{\operatorname{original}} \sim p_{\theta}(x | z_{\operatorname{original}}),\;
\hat{X}_{\operatorname{shifted}} \sim p_{\theta}(x | z_{\operatorname{shifted}}),    
\end{equation}
we propose the following topological loss term (Algorithm \ref{alg:tdloss}):
\begin{equation}
\label{eq:td_def}
    \mathcal{L}_{TD} = \operatorname{RTD}^{(p)}(\hat{X}_{\operatorname{original}}, \hat{X}_{\operatorname{shifted}}),
\end{equation}
where the superscript $(p)$ in $\operatorname{RTD}^{(p)}$ stands for using sum of the lengths of intervals in R-Cross-Barcode$_1$ to the $p-$th power. 
The $\mathcal{L}_{TD}$ term imposes a penalty for data point clouds having different topological structures, like the 2nd and the 3rd rows in Figure \ref{fig:rtd_traverse}. Both standard values $p=1$ and $p=2$ perform well. In some circumstances, the $p = 2$ value is more appropriate because it penalizes significant variations in topology structures. 

In this work, we propose to use the topological loss term $\mathcal{L}_{TD}$, in addition to the VAE-based loss:
\begin{equation}
\label{eq:objective}
\mathcal{L}=\mathcal{L}_{VAE-based} + \gamma\mathcal{L}_{TD}. 
\end{equation}

All variants of VAEs (classical VAE, $\beta$-VAE, FactorVAE, $\beta$-TCVAE, ControlVAE, DAVA) are modified accordingly.
The computational complexity of $\mathcal{L}_{TD}$ is discussed in Appendix \ref{app:complexity}. We analyze the sensitivity of the proposed approach on the value of $\gamma$ in (\ref{eq:objective}) in Appendix \ref{app:sensitivity}.

\subsection{Gradient Orthogonalization}
As with all regularization terms, the $\mathcal{L}_{TD}$  minimization may lead to a lack of reconstruction quality. In order to achieve state-of-the-art results while minimizing the topological regularization term $\mathcal{L}_{TD}$, we apply the gradient orthogonalization between $\mathcal{L}_{TD}$ and the reconstruction loss term $\mathcal{L}_{rec}$. Specifically, if the scalar product between $\nabla_{\phi,\theta} \mathcal{L}_{rec}$ and $\nabla_{\phi,\theta}\mathcal{L}_{TD}$ is negative, then we adjust the gradients from our $\mathcal{L}_{TD}$ loss to be orthogonal to those from $\mathcal{L}_{rec}$ by applying the appropriate linear transformation:
\begin{equation}
\nabla^{ort}\mathcal{L}_{TD} = \nabla\mathcal{L}_{TD} - \frac{\langle\nabla\mathcal{L}_{TD}, \nabla \mathcal{L}_{rec}\rangle}{\langle\nabla\mathcal{L}_{rec}, \nabla \mathcal{L}_{rec}\rangle}\nabla\mathcal{L}_{rec}.
\end{equation}
This technique helps to maintain a balance between the reconstruction quality and the topological regularization, thus resulting in improved overall performance. This follows essentially from the properties of the gradient of a differentiable function: making a step of length $\delta$ in a direction orthogonal to the gradient does not change the function value up to higher order $\mathcal{O}(\delta^2)$, while moving in a direction that has negative scalar product with the gradient decreases the function value. We provide an ablation study of gradient orthogonalization technique in Appendix \ref{app:ablation}.

\begin{table*}
\centering
\setlength\extrarowheight{-1pt}
\setlength\tabcolsep{2.25pt}
\caption{Evaluation on the benchmark datasets. \textbf{Bold} denotes the best variant in the pair with vs. without the TopDis loss. {\textcolor{blue} Blue} denotes the best method for a dataset/metric.}
\vspace{10pt}
\label{tbl:results}
\begin{tabularx}{\textwidth}{
X
>{\hsize=0.5\hsize}X
>{\hsize=0.5\hsize}X
>{\hsize=0.5\hsize}X
>{\hsize=0.5\hsize}X
}
\toprule
\textbf{Method}                    & \textbf{FactorVAE score}        & \textbf{MIG}                        & \textbf{SAP}                        & \textbf{DCI, dis.}       \\ \toprule
\multicolumn{5}{c}{dSprites}                                                                                                              \\ \toprule
$\beta$-VAE & $0.807 \pm 0.037$ & $0.272 \pm 0.101$ & $0.065 \pm 0.002$ & $0.440 \pm 0.102$ \\
$\beta$-VAE + TopDis (ours) & \textcolor{blue}{$\mathbf{0.833 \pm 0.016}$} & $\mathbf{0.348 \pm 0.028}$ & $\mathbf{0.066 \pm 0.015}$ & $\mathbf{0.506 \pm 0.050}$ \\ \midrule 

FactorVAE  & $0.819 \pm 0.028$   & $0.295 \pm 0.049$          & $0.053 \pm 0.006$ & $\mathbf{0.534 \pm 0.029}$ \\ 
FactorVAE + TopDis (ours) & $\mathbf{0.824 \pm 0.038}$ & \textcolor{blue}{$\mathbf{0.356 \pm 0.025}$} & \textcolor{blue}{$\mathbf{0.082 \pm 0.001}$} & $0.521 \pm 0.044$ \\ \midrule 

$\beta$-TCVAE & $0.810 \pm 0.058$ & $0.332 \pm 0.029$ & $0.045 \pm 0.004$ & $0.543 \pm 0.049$ \\
$\beta$-TCVAE + TopDis (ours) & $\mathbf{0.821 \pm 0.034}$ & $\mathbf{0.341 \pm 0.021}$ & $\mathbf{0.051 \pm 0.004}$ & $\mathbf{0.556 \pm 0.042}$ \\ \midrule 

ControlVAE & $0.806 \pm 0.012$ & $0.333 \pm 0.037$ & $0.056 \pm 0.002$ & $0.557 \pm 0.009$ \\
ControlVAE + TopDis (ours) & $\mathbf{0.810 \pm 0.012}$ & $\mathbf{0.344 \pm 0.029}$ & $\mathbf{0.059 \pm 0.002}$ & \textcolor{blue}{$\mathbf{0.578 \pm 0.007}$} \\ \midrule 

DAVA & $0.746 \pm 0.099$ & $0.253 \pm 0.058$ & $0.024 \pm 0.015$ & $0.395 \pm 0.054$ \\
DAVA + TopDis (ours) & $\mathbf{0.807 \pm 0.010}$ & $\mathbf{0.344 \pm 0.010}$ & $\mathbf{0.048 \pm 0.012}$ & $\mathbf{0.551 \pm 0.019}$ \\
\toprule
\multicolumn{5}{c}{3D Shapes}                                                                                                              \\ \toprule
$\beta$-VAE & $0.965 \pm 0.060$ & $0.740 \pm 0.141$ & $0.143 \pm 0.071$ & $0.913 \pm 0.147$ \\
$\beta$-VAE + TopDis (ours) & \textcolor{blue}{$\mathbf{1.0 \pm 0.0}$} & \textcolor{blue}{$\mathbf{0.839 \pm 0.077}$} & \textcolor{blue}{$\mathbf{0.195 \pm 0.030}$} & \textcolor{blue}{$\mathbf{0.998 \pm 0.004}$} \\ \midrule

FactorVAE & $0.934 \pm 0.058$ & $0.698 \pm 0.151$ & $0.099 \pm 0.064$ & $0.848 \pm 0.129$\\ 
FactorVAE + TopDis (ours) & $\mathbf{0.975 \pm 0.044}$ & $\mathbf{0.779 \pm 0.036}$ & $\mathbf{0.159 \pm 0.032}$ & $\mathbf{0.940 \pm 0.089}$\\ \midrule

$\beta$-TCVAE & $0.909 \pm 0.079$ & $0.693 \pm 0.053$ & $0.113 \pm 0.070$ & $0.877 \pm 0.018$\\ 
$\beta$-TCVAE + TopDis (ours) & \textcolor{blue}{$\mathbf{1.0 \pm 0.0}$} & $\mathbf{0.751 \pm 0.051}$ & $\mathbf{0.147 \pm 0.064}$ & $\mathbf{0.901 \pm 0.014}$\\ \midrule

ControlVAE & $0.746 \pm 0.094$ & $0.433 \pm 0.094$ & $0.091 \pm 0.068$ & $0.633 \pm 0.093$\\
ControlVAE + TopDis (ours) & $\mathbf{0.806 \pm 0.046}$ & $\mathbf{0.591 \pm 0.055}$ & $\mathbf{0.125 \pm 0.02}$ & $\mathbf{0.795 \pm 0.098}$\\ \midrule

DAVA & $0.800 \pm 0.095$ & $0.625 \pm 0.061$ & $0.099 \pm 0.016$ & $0.762 \pm 0.088$ \\
DAVA + TopDis (ours) & $\mathbf{0.847 \pm 0.092}$ & $\mathbf{0.679 \pm 0.112}$ & $\mathbf{0.101 \pm 0.043}$ & $\mathbf{0.836 \pm 0.074}$ \\
\toprule
\multicolumn{5}{c}{3D Faces}                                                                                                              \\ \toprule
$\beta$-VAE & \textcolor{blue}{$\mathbf{1.0 \pm 0.0}$} & $\mathbf{0.561 \pm 0.017}$ & $\mathbf{0.058 \pm 0.008}$ & \textcolor{blue}{$\mathbf{0.873 \pm 0.018}$} \\
$\beta$-VAE + TopDis (ours) & \textcolor{blue}{$\mathbf{1.0 \pm 0.0}$} & $0.545 \pm 0.005$ & $0.052 \pm 0.004$ & $0.854 \pm 0.013$ \\ \midrule

FactorVAE & \textcolor{blue}{$\mathbf{1.0 \pm 0.0}$} & $0.593 \pm 0.058$ & $0.061 \pm 0.014$ & $0.848 \pm 0.011$ \\
FactorVAE + TopDis (ours) & \textcolor{blue}{$\mathbf{1.0 \pm 0.0}$} & \textcolor{blue}{$\mathbf{0.626 \pm 0.026}$} & $\mathbf{0.062 \pm 0.013}$ & $\mathbf{0.867 \pm 0.037}$ \\ \midrule

$\beta$-TCVAE & \textcolor{blue}{$\mathbf{1.0 \pm 0.0}$} & $0.568 \pm 0.063$ & $0.060 \pm 0.017$ & $0.822 \pm 0.033$ \\
$\beta$-TCVAE + TopDis (ours) & \textcolor{blue}{$\mathbf{1.0 \pm 0.0}$} & $\mathbf{0.591 \pm 0.058}$ & $\mathbf{0.062 \pm 0.011}$ & $\mathbf{0.859 \pm 0.031}$ \\ \midrule

ControlVAE & \textcolor{blue}{$\mathbf{1.0 \pm 0.0}$} & $0.447 \pm 0.011$ & $0.058 \pm 0.008$ & $0.713 \pm 0.007$ \\
ControlVAE + TopDis (ours) & \textcolor{blue}{$\mathbf{1.0 \pm 0.0}$} & $\mathbf{0.477 \pm 0.004}$ & \textcolor{blue}{$\mathbf{0.074 \pm 0.007}$} & $\mathbf{0.760 \pm 0.014}$  \\ \midrule

DAVA & \textcolor{blue}{$\mathbf{1.0 \pm 0.0}$} & $0.527 \pm 0.002$ & $0.047 \pm 0.009$ & $\mathbf{0.822 \pm 0.006}$ \\
DAVA + TopDis (ours) & \textcolor{blue}{$\mathbf{1.0 \pm 0.0}$} & $\mathbf{0.536 \pm 0.012}$ & $\mathbf{0.052 \pm 0.011}$ & $0.814 \pm 0.008$
\\
\toprule
\multicolumn{5}{c}{MPI 3D}                                                                                                              \\ \toprule
$\beta$-VAE & $0.428 \pm 0.054$ & $0.221 \pm 0.087$ & $0.092 \pm 0.035$ & $0.238 \pm 0.049$ \\
$\beta$-VAE + TopDis (ours) & $\mathbf{0.479 \pm 0.040}$ & $\mathbf{0.335 \pm 0.056}$ & $\mathbf{0.172 \pm 0.032}$ & $\mathbf{0.337 \pm 0.036}$ \\ \midrule

FactorVAE & $0.589 \pm 0.053$ & $0.336 \pm 0.056$ & $0.179 \pm 0.052$ & $0.391 \pm 0.056$ \\ 
FactorVAE + TopDis (ours) & \textcolor{blue}{$\mathbf{0.665 \pm 0.041}$} & \textcolor{blue}{$\mathbf{0.377 \pm 0.053}$} & \textcolor{blue}{$\mathbf{0.238 \pm 0.040}$} & \textcolor{blue}{$\mathbf{0.438 \pm 0.065}$} \\ \midrule

$\beta$-TCVAE & $0.377\pm0.039$ & $0.168\pm0.021$ & $0.084\pm0.012$ & $0.233\pm0.059$\\ 
$\beta$-TCVAE + TopDis (ours) & $\mathbf{0.501 \pm 0.023}$  & $\mathbf{0.287 \pm 0.011}$ & $\mathbf{0.149 \pm 0.006}$ & $\mathbf{0.356 \pm 0.045}$\\ \midrule

ControlVAE & $0.391 \pm 0.021$ & $0.180 \pm 0.048$ & $0.107 \pm 0.003$ & $0.178 \pm 0.037$ \\
ControlVAE + TopDis (ours) & $\mathbf{0.554 \pm 0.026}$ & $\mathbf{0.232 \pm 0.016}$ & $\mathbf{0.154 \pm 0.003}$ & $\mathbf{0.274 \pm 0.028}$ \\ \midrule

DAVA & $0.404 \pm 0.080$ & $0.234 \pm 0.075$ & $0.086 \pm 0.043$ & $0.268 \pm 0.051$ \\
DAVA + TopDis (ours) & $\mathbf{0.606 \pm 0.036}$ & $\mathbf{0.337 \pm 0.067}$ & $\mathbf{0.181 \pm 0.041}$ & $\mathbf{0.401 \pm 0.049}$ \\
\bottomrule
\end{tabularx}
\end{table*}

\section{Experiments}
\subsection{Experiments on Standard Benchmarks}
\vskip-.1in
In the experimental section of our work, we evaluate the effectiveness of the proposed topology-based loss.
Specifically, we conduct a thorough analysis of the ability of our method to learn disentangled latent spaces using various datasets and evaluation metrics. We compare the results obtained by our method with the state-of-the-art models and demonstrate the advantage of our approach in terms of disentanglement and reconstruction quality.

\textbf{Datasets}.
We used popular benchmarks: dSprites \citep{dsprites17}, 3D Shapes \citep{3dshapes18}, 3D Faces \citep{paysan20093d}, MPI 3D \citep{mpi3d}, CelebA \citep{liu2015deep}. See the description of the datasets in Appendix \ref{app:datasets}.
Although the datasets dSprites, 3D Shapes, 3D Faces are synthetic, the known true factors of variation allow accurate supervised evaluation of disentanglement. Hence, these datasets are commonly used in both classical and most recent works on disentanglement \citep{burgess2017understanding, kim2018disentangling, estermann2023dava, roth2022disentanglement}.
Finally, we examine the real-life setup with the CelebA \citep{liu2015deep} dataset.

\textbf{Methods}.
To demonstrate that the proposed TopDis loss contributes to learning disentangled representations, we combine it with classical VAE \citep{kingma2013auto}, $\beta$-VAE \citep{higgins2017beta}, FactorVAE \citep{kim2018disentangling}, $\beta$-TCVAE \citep{chen2018isolating}, ControlVAE \citep{shao2020controlvae} and DAVA \citep{estermann2023dava}.
Following the previous work \citep{kim2018disentangling}, we used similar architectures for the encoder, decoder and discriminator (see Appendix \ref{app:architectures}), the same for all models. The hyperparameters and other training details are in Appendix \ref{app:hyperparams}. We set the latent space dimensionality to $10$. Since the quality of disentanglement has high variance w.r.t. network initialization \citep{locatello2019challenging}, we conducted multiple runs of our experiments using different initialization seeds (see Appendix \ref{sec:mann-witney_crit}) and averaged results.

\begin{figure*}[t]
\centering
\begin{subfigure}{0.49\textwidth}
  \centering
  \includegraphics[width=1\linewidth]{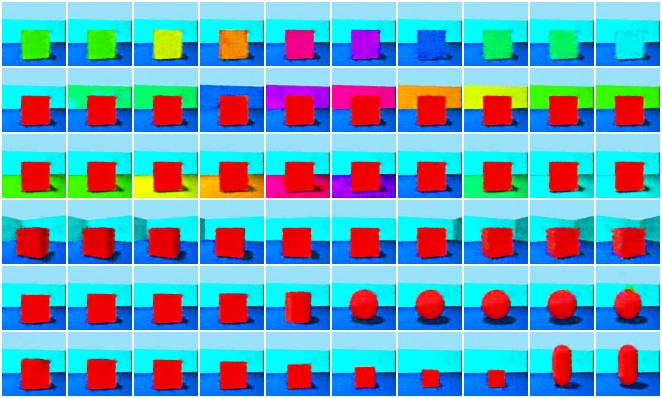}
\label{fig:FactorVAE_3dshapes_traverse_main_text}
\end{subfigure}
\hfill
\begin{subfigure}{0.49\textwidth}
  \centering
  \includegraphics[width=1\linewidth]{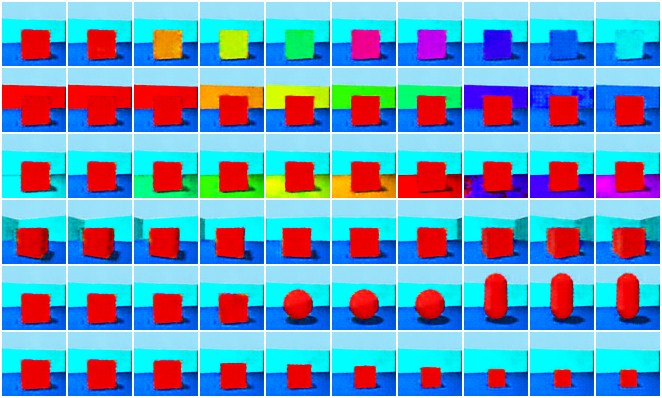}
\label{fig:FactorVAE_RTD_3dshapes_traverse_main_text}
\end{subfigure}
\vspace{-0.75cm}
\caption{FactorVAE (left) and FactorVAE + TopDis (right) latent traversals on 3D Shapes.}
\label{fig:latent_traversals_3d_shapes}
\vskip-.2in
\end{figure*}

\textbf{Evaluation}. Not all existing metrics were shown to be equally useful and suitable for disentanglement \citep{dittadi2021onthetransfer, locatello2019challenging}. Due to this, hyperparameter tuning and model selection may become controversial. Moreover, in the work \citet{zaidi2020measuring}, the authors conclude that the most appropriate metric is DCI disentanglement score \citep{eastwood2018framework}, the conclusion which coincides with another line of research \citet{roth2022disentanglement}. Based on the existing results about metrics' applicability, we restricted evaluation to measuring the following disentanglement metrics: the Mutual Information Gap (MIG) \citep{chen2018isolating}, the FactorVAE score \citep{kim2018disentangling}, DCI disentanglement score, and Separated Attribute Predictability (SAP) score \citep{kumar2017variational}. Besides its popularity, these metrics cover all main approaches to evaluate the disentanglement of generative models \citep{zaidi2020measuring}: information-based (MIG), predictor-based (SAP score, DCI disentanglement score), and intervention-based (FactorVAE score).

\subsubsection{Quantitative Evaluation}
\vskip-.1in

First of all, we study the TopDis loss as a self-sufficient disentanglement objective by adding it to VAE. Table \ref{tbl:vae_topdis_results} in Appendix \ref{app:vae_topdis} shows that disentanglement metrics are improved for all the datasets. 
Next, we add the TopDiss loss to state-of-the-art models.
As demonstrated in Table \ref{tbl:results}, the models trained with the auxiliary TopDis loss outperform the original ones for all datasets and almost all quality measures.
The addition of the TopDis loss improves the results as evaluated by FactorVAE score, MIG, SAP, DCI: on dSprites up to +8\%, +35\%, +100\%, +39\%, on 3D Shapes up to +8\%, +36\%, +60\%, +25\%, on 3D Faces up to +6\%, +27\%, +6\% and up to +50\%, +70\%, +110\%, +53\% on MPI 3D respectively across all models.
The best variant for a dataset/metrics is almost always a variant with the TopDis loss, in 94\% cases.
In addition, our approach preserves the reconstruction quality, see \mbox{Table \ref{tab:reconstruction_all}}, Appendix \ref{app:reconstruction}.
\subsubsection{Qualitative Evaluation}
\vskip-.1in
In order to qualitatively evaluate the ability of our proposed TopDis loss to learn disentangled latent representations, we plot the traversals along a subset of latent codes that exhibit the most significant changes in an image. As a measure of disentanglement, it is desirable for each latent code to produce a single factor of variation. We compare traversals from FactorVAE and FactorVAE+TopDis decoders. The corresponding Figures \ref{fig:latent_traversals}, \ref{fig:celeba_main} and a detailed discussion are in Appendix \ref{app:traversals}.

For the \textbf{dSprites} dataset, the simple FactorVAE model has entangled rotation and shift factors, while in FactorVAE+TopDis these factors are isolated. For the \textbf{3D Shapes} (Figure \ref{fig:latent_traversals_3d_shapes}),  FactorVAE+TopDis learns disentangled shape and scale factors, while classical FactorVAE doesn't.
In \textbf{3D Faces}, FactorVAE+TopDis better disentangles azimuth, elevation, and lighting. Especially for lighting, facial attributes such as the chin, nose, and eyes are preserved for the ``lightning'' axis traversal.
For \textbf{MPI 3D}, FactorVAE+TopDis successfully disentangles size and elevation factors.
Finally, for \textbf{CelebA}, FactorVAE+TopDis disentangles skin tone and lightning, while in FactorVAE they are entangled with background and hairstyle. 

\subsection{Learning Disentangled Representations from Correlated Data}
\vskip-.1in
Existing methods for disentanglement learning make unrealistic assumptions about statistical independence of factors of variations
\citep{trauble2021disentangled}. Synthetic datasets (dSprites, 3D Shapes, 3D Faces, MPI 3D) also share this assumption. However, in the real world, causal factors are typically correlated. We carry out a series of experiments with shared confounders (one factor correlated to all others, \citep{roth2022disentanglement}). The TopDis loss isn't based on assumptions of statistical independence. The addition of the TopDis loss gives a consistent improvement in all quality measures in this setting, see Table \ref{tbl:corr_factors_results} in 
Appendix \ref{app:correlated_factors}.

\subsection{Unsupervised Discovery of Disentangled Directions in StyleGAN}
\vskip-.1in
\begin{figure}[tbp]
\centering
\includegraphics[width=0.45\textwidth]{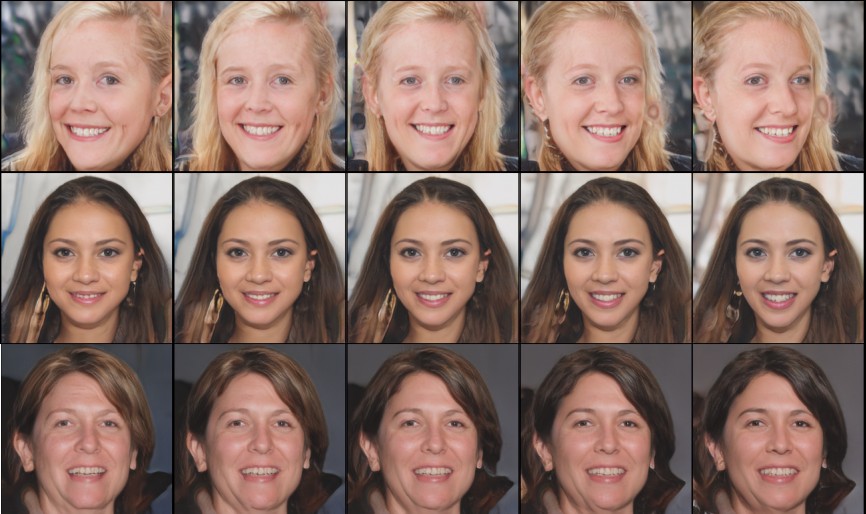}
\vspace{-0.25cm}
\caption{Three disentangled directions discovered by TopDis in StyleGAN: azimuth, smile, and hair color.}
\vskip-.25in
\label{fig:stylegan_teaser}
\end{figure}
We perform additional experiments to study the ability of the proposed topology-based loss to infer disentangled directions in a pretrained StyleGAN \citep{karras2019style}. We searched for disentangled directions within the space of principal components in latent space by optimizing the multi-scale topological  difference after a shift along this axis
$\operatorname{RTD}(\hat{X}_{\operatorname{original}}, \hat{X}_{\operatorname{shifted}})$. We were able to find three disentangled directions: azimuth, smile, hair color. 
See Figure \ref{fig:stylegan_teaser} and Appendix \ref{app:stylgan} for more details.
Comparison of methods dedicated to the unsupervised discovery of disentangled directions in StyleGAN is qualitative since the FFHQ dataset doesn't have labels. We do not claim that our method outperforms alternatives \citep{harkonen2020ganspace}, as our goal is rather to demonstrate the applicability of the TopDis loss for this problem.

\section{Conclusion}
\vskip-0.1in
Our method, the Topological Disentanglement, has demonstrated its effectiveness in learning disentangled representations, in an unsupervised manner. The experiments on the dSprites, 3D Shapes, 3D Faces, and MPI 3D datasets have shown that an addition of the proposed TopDis loss improves VAE, $\beta$-VAE, FactorVAE, $\beta$-TCVAE, ControlVAE, and DAVA models in terms of disentanglement scores (MIG, FactorVAE, SAP, DCI disentanglement) while preserving the reconstruction quality. 
Inside our method, there is the idea of applying the topological dissimilarity to optimize disentanglement that can be added to any existing approach or used alone. 
We proposed to apply group(oid) action shifts preserving the Gaussian distribution in the latent space. To preserve the reconstruction quality, the gradient orthogonalization was used. Our method isn't based on the statistical independence assumption and brings improvement in quality measures even if factors of variation are correlated.
In this paper, we limited ourselves to the image domain for easy visualization of disentangled directions.
Extension to other domains (robotics, time series, etc.) is an interesting avenue for further research.

\section*{Acknowledgements}
The work was supported by the Analytical center under the RF Government (subsidy agreement 000000D730321P5Q0002, Grant No. 70-2021-00145 02.11.2021).

\section*{Impact Statement}
Disentanglement learning will lead to better generative models in various domains like image, video, text, etc. Generative models have a high potential industrial and societal impact since they may lead to creation of multi-modal chat bots, AI-assisted image and video production, etc. The danger of deepfakes should be taken into account.

\bibliography{icml2024}
\bibliographystyle{icml2024}

\newpage
\appendix
\onecolumn

\section{Datasets}
\label{app:datasets}

This section provides a brief overview of the benchmark datasets along with sample images.

\begin{figure}[!ht]
\centering
\begin{minipage}{0.5\textwidth}
  \centering
  \includegraphics[width=0.9\linewidth]{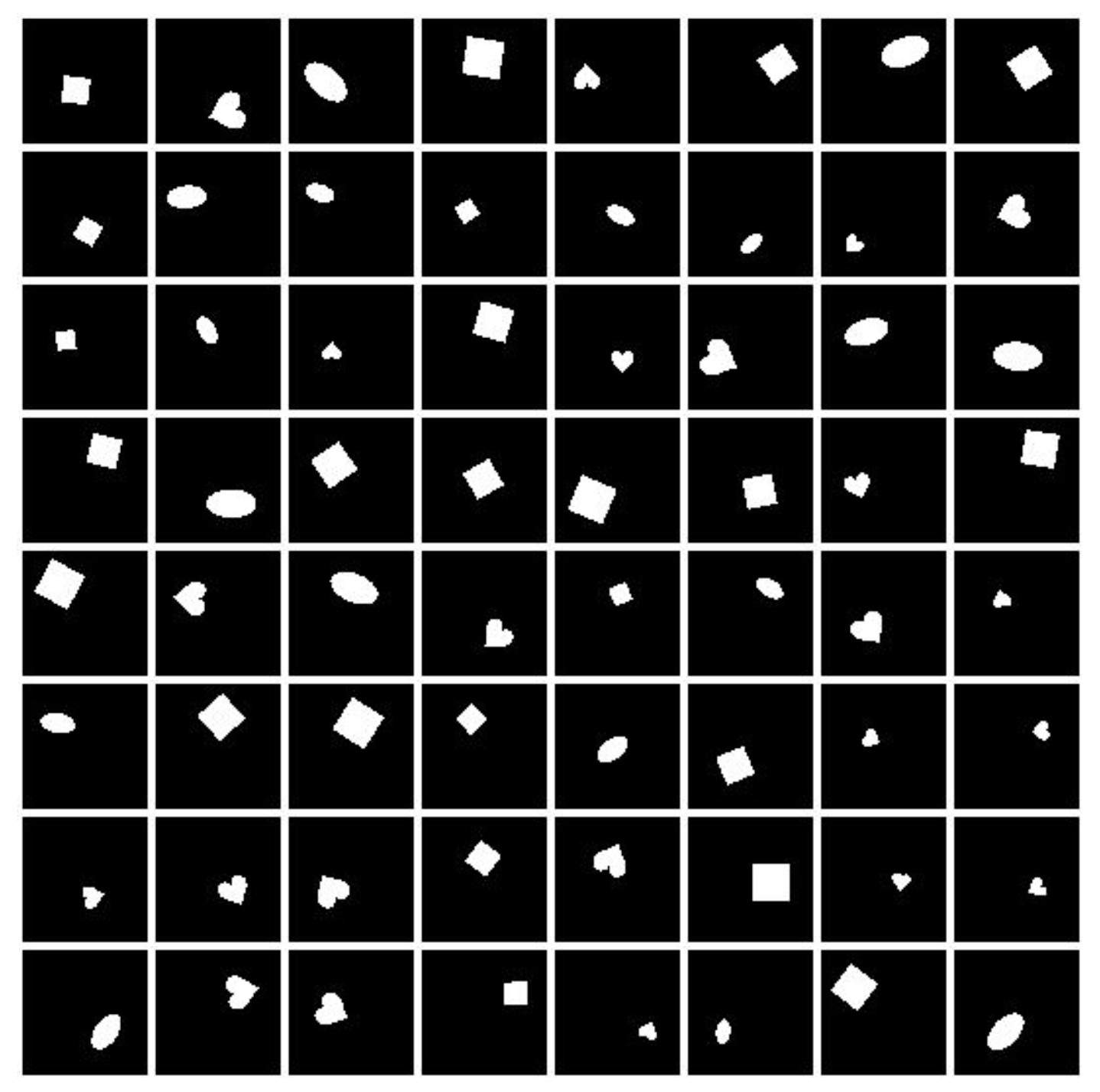}
  \captionof{figure}{Samples from dSprites.}
  \label{fig:2d_sample}
\end{minipage}%
\begin{minipage}{0.5\textwidth}
  \centering
  \includegraphics[width=0.9\linewidth]{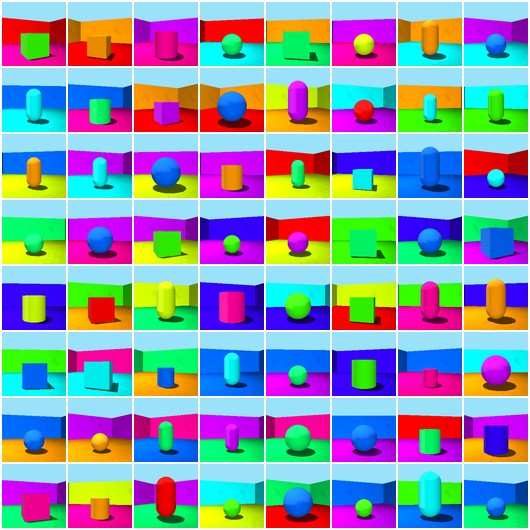}
  \captionof{figure}{Samples from 3D Shapes.}
  \label{fig:3d_sample}
\end{minipage}
\end{figure}

\textbf{dSprites} contains 2D shapes generated procedurally from five independent latent factors: shape, scale, rotation, x-coordinate, and y-coordinate of a sprite. See Figure \ref{fig:2d_sample} for sample images. 

\textbf{3D Shapes} consists of 3D scenes with six generative factors: floor hue, wall hue, orientation, shape, scale, and shape color. See Figure \ref{fig:3d_sample} for sample images.

\begin{figure}[!ht]
\centering
\begin{minipage}{0.5\textwidth}
  \centering
  \includegraphics[width=0.9\linewidth]{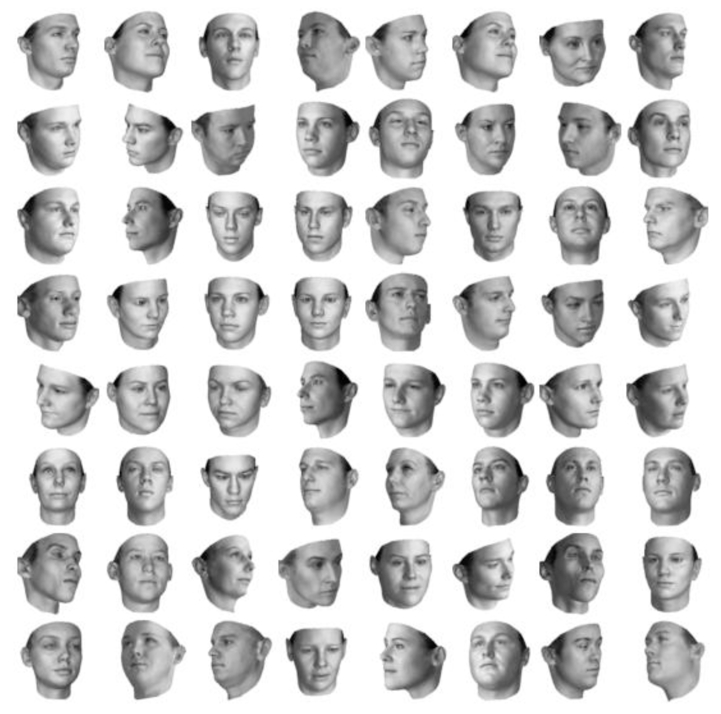}
  \captionof{figure}{Samples from 3D Faces.}
  \label{fig:3dfaces_sample}
\end{minipage}%
\begin{minipage}{0.5\textwidth}
  \centering
  \includegraphics[width=0.9\linewidth]{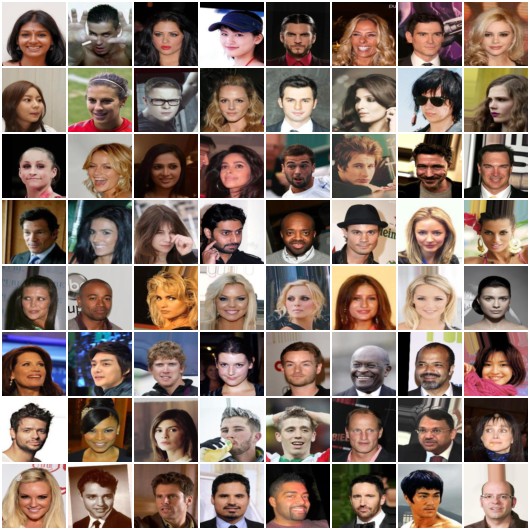}
  \captionof{figure}{Samples from CelebA.}
  \label{fig:celeba_sample}
\end{minipage}
\end{figure}

\textbf{3D Faces} The 3D Faces dataset consists of 3D rendered faces with four generative factors: face id, azimuth, elevation, lighting. See Figure \ref{fig:3dfaces_sample} for sample images.

\begin{figure}[!ht]
\centering
\begin{minipage}{0.5\textwidth}
  \centering
  \includegraphics[width=0.9\linewidth]{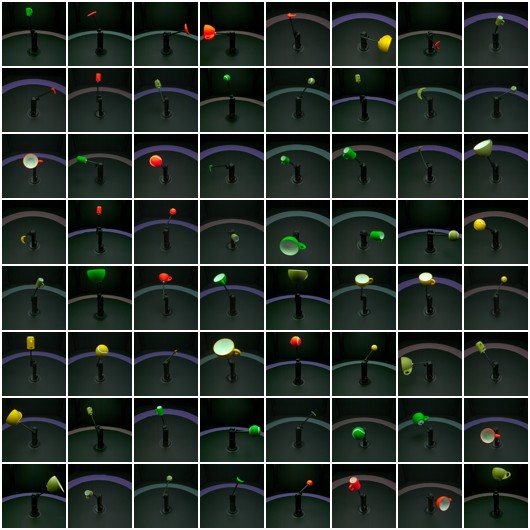}
  \captionof{figure}{Samples from MPI 3D.}
  \label{fig:mpi3d_sample}
\end{minipage}%
\begin{minipage}{0.5\textwidth}
  \centering
  \includegraphics[width=0.9\linewidth]{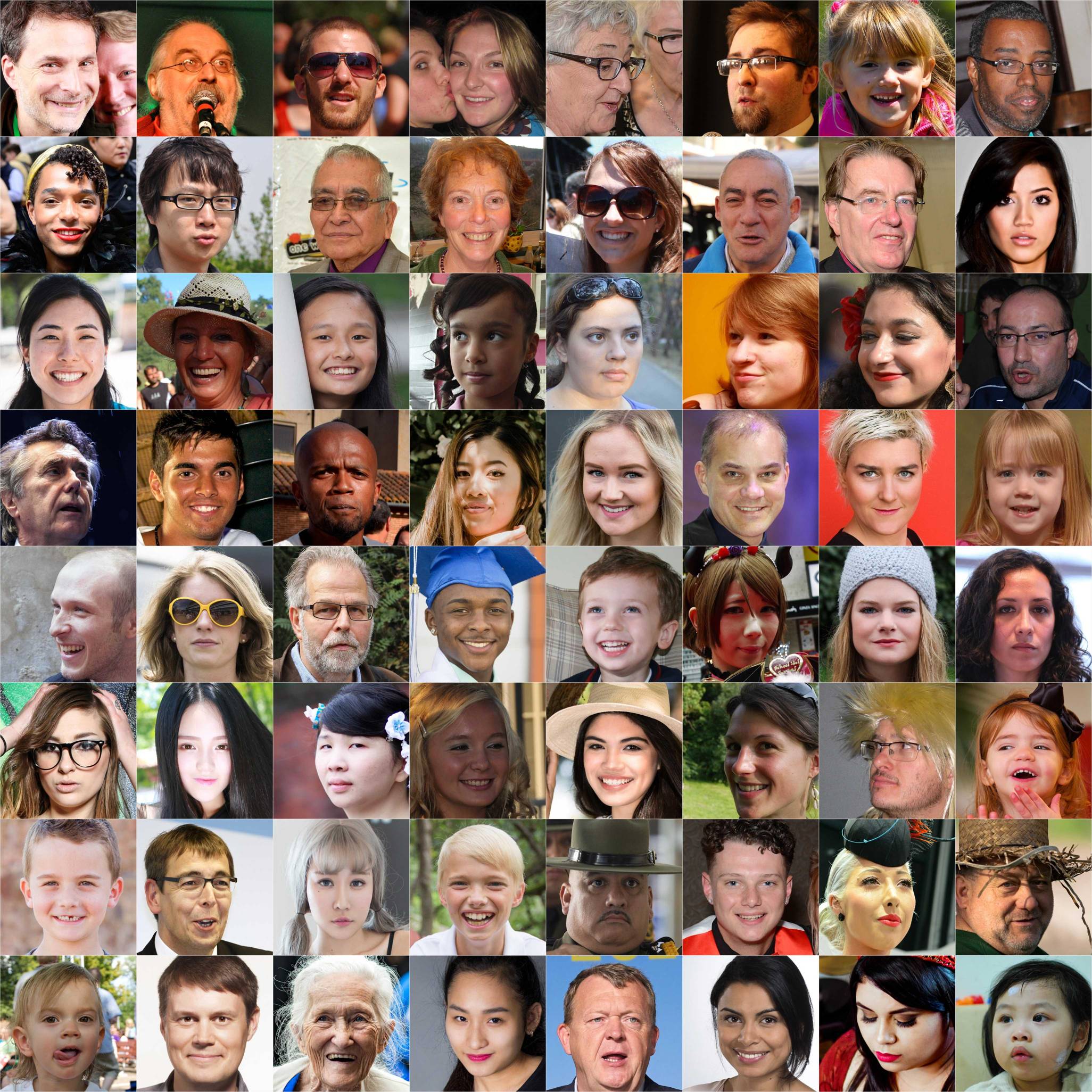}
  \captionof{figure}{Samples from FFHQ.}
  \label{fig:ffhq_sample}
\end{minipage}
\end{figure}

\textbf{MPI 3D} contains images of physical 3D objects with seven generative factors: color, shape, size, camera height, background color, horizontal and vertical axes. We used the MPI3D-Complex version which provides samples of complex real-world shapes from the robotic platform.
See Figure \ref{fig:mpi3d_sample} for sample images.

\textbf{CelebA} provides images of aligned faces of celebrities. This dataset doesn't have any ground truth generative factors because of its real-world nature. Figure \ref{fig:3dfaces_sample} demonstrates the sample images.

\textbf{FFHQ} contains high-quality images of human faces at resolution. Similarly to CelebA, due to its real-world origin, this dataset doesn't have any ground truth factors of variation. Figure \ref{fig:ffhq_sample} provides the sample images.

In our work, we used both synthetic and real-world datasets as in recent state-of-the-art research in disentanglement, see  \citet{roth2022disentanglement, estermann2023dava, shao2020controlvae, chen2018isolating, kim2018disentangling}. 
\citet{estermann2023dava}
\citet{shao2020controlvae}
\citet{chen2018isolating}
\citet{kim2018disentangling} 
We highlight that we utilize the MPI3D-Real-Complex version of the MPI 3D dataset, which was developed based on robotic platform in real-world setting and contains complex real-world shapes.

\section{Proof of  Proposition \ref{prop:groupact} }\label{proof:groupact}

\textit{Proof.}
a) Two consecutive shifts defined in equation (\ref{shifts}) give
\begin{multline*}
   F^{-1}(F( F^{-1}(F(z \mid \rho,\sigma^2) + C_1 \mid \rho,\sigma^2)+C_2\mid \rho,\sigma^2)= F^{-1}(F(z \mid \rho,\sigma^2) + C_1+C_2 \mid \rho,\sigma^2) 
\end{multline*} So the two consecutive shifts with $C_1,C_2$ is the same as the single shift with $C_1+C_2$.

b) We have for a given shift with parameter $C$ and any pair of shifted points $z_{\operatorname{shifted}}, \tilde{z}_{\operatorname{shifted}}\in \mathbb{R}$ : 
\begin{equation}
    F(\tilde{z}_{\operatorname{shifted}})-F(z_{\operatorname{shifted}})=(F(\tilde{z})+C)-(F(z)+C)=F(\tilde{z})-F(z)
\end{equation}i.e. if the shift of  points $z, \tilde{z}\in \mathbb{R}$ is defined, then the ${N}(\rho, \sigma^2)$ measure of the line segment $[z, \tilde{z}]$ is preserved under the shift. 

c) Conversely, if for  $z, \tilde{z}\in \mathbb{R}$  the ${N}(\rho, \sigma^2)$ measure of the line segment $[z, \tilde{z}]$ is preserved under the shift, i.e.
$
    F(\tilde{z}_{\operatorname{shifted}})-F(z_{\operatorname{shifted}})=F(\tilde{z})-F(z)$, then  setting $z$ to any fixed value, for example $z=-\infty$, we get 
$F(\tilde{z}_{\operatorname{shifted}})=F(\tilde{z})+C$.
\qed

Notice also that $F(z_{\operatorname{shifted}})-F(z)=C$, so the three orange curvilinear rectangles on Figure \ref{fig:probshift} have the same area $C=1/8$.

Recall that $F(z\mid\rho, \sigma^2) = \frac{1}{\sigma\sqrt{2\pi}} \int_{-\infty}^z \exp \left( -\frac{(t - \rho)^2}{2\sigma^2} \right) dt$ denotes here the cumulative function of the Gaussian distribution ${N}(\rho, \sigma^2)$. 

\begin{remark}
    Notice that, during the calculation of the topological term, we do not consider the data points with $F(z)+C>1$ ($F(z)+C<0$), i.e. whose latent codes are already at the very right (left) tail  of the distribution and which thus cannot be shifted to the right (respectfully, left). 
\end{remark}

\section{Proof of Proposition \ref{prop:qzi}} \label{proof:qzi}
\textit{Proof.}
a) The shift defined by (\ref{shifts}) for the distribution $q(z_i)$ acting on the latent space, preserves also any $q(z_j)$ for $ j\neq i$. b) The result follows from the case of an arbitrary distribution over a pair of random variables $z_1, z_2$. For two variables, it follows from the Bayes formula that the shifts of $z_1$ preserve the conditional $q(z_2\vert z_1)$. Since the group(oid) action is transitive it follows that the conditional does not depend on $z_1$, and hence $q(z_1,z_2)=q(z_1)q(z_2)$. 

\section{Architecture Details}
\label{app:architectures}
Table \ref{tab:encoder_decoder} demonstrates the architecture of VAE model while
the discriminator's architecture is presented in Table \ref{tab:discriminator}.
In the experiments with VAE(+TopDis), $\beta$-VAE(+TopDis), FactorVAE(+TopDis), $\beta$-TCVAE(+TopDis), ControlVAE(+TopDis), DAVA(+TopDis) (Tables \ref{tbl:vae_topdis_results}, \ref{tbl:results}, \ref{tab:reconstruction_all}), we used the following architecture configurations: 
\begin{itemize}
\item dSprites: $\text{num channels}=1, m_1=2, m_2=2, m_3=4, m_4=4, n=5$;
\item 3D Shapes: $\text{num channels}=3, m_1=1, m_2=1, m_3=1, m_4=2, n=5$;
\item 3D Faces: $\text{num channels}=1, m_1=1, m_2=1, m_3=1, m_4=2, n=5$;
\item MPI 3D: $\text{num channels}=3, m_1=1, m_2=1, m_3=1, m_4=2, n=6$;
\item CelebA: $\text{num channels}=3, m_1=1, m_2=1, m_3=1, m_4=2, n=5$.
\end{itemize}

\begin{table*}[!ht]
\begin{center}
     \caption{Encoder and Decoder architecture for the dSprites experiments.}
    \begin{tabular}{l l c c}
    \toprule
        Encoder & Decoder \\
        \midrule
        Input: $64 \times 64 \times \text{num channels}$ & Input: $\mathbb{R}^{10}$ \\
        $4 \times 4$ conv, $32$ ReLU, stride $2$ & $1 \times 1$ conv, $128 \times \text{m}_4$ ReLU, stride $1$ \\
        $4 \times 4$ conv, $32 \cdot \text{m}_1$ ReLU, stride $2$ & $4 \times 4$ upconv, $64 \cdot \text{m}_3$ ReLU, stride $1$ \\
        $4 \times 4$ conv, $64 \cdot \text{m}_2$ ReLU, stride $2$ & $4 \times 4$ upconv, $64 \cdot \text{m}_2$ ReLU, stride $2$ \\
        $4 \times 4$ conv, $64 \cdot \text{m}_3$ ReLU, stride $2$ & $4 \times 4$ upconv, $32 \cdot \text{m}_1$ ReLU, stride $2$ \\
        $4 \times 4$ conv, $128 \cdot \text{m}_4$ ReLU, stride $1$ & $4 \times 4$ upconv, $32$ ReLU, stride $2$ \\
        $1 \times 1$ conv, $2 \times 10$, stride $1$ & $4 \times 4$ upconv, $1$, stride $2$ \\
        \bottomrule
        \end{tabular}
    \label{tab:encoder_decoder}
\end{center}
\end{table*}

\begin{table*}[!ht]
\begin{center}
     \caption{FactorVAE Discriminator architecture.}
    \begin{tabular}{l c}
        \toprule
        Discriminator \\
        \midrule
        $\left[\text{FC, }1000\text{ leaky ReLU }\right] \times n$\\
        FC, $2$ \\
        \bottomrule
        \end{tabular}
    \label{tab:discriminator}
\end{center}
\end{table*}

\section{Reconstruction Error}
\label{app:reconstruction}

We provide the reconstruction error for all the evaluated models in Table \ref{tab:reconstruction_all}.

\begin{table}
\begin{center}
\setlength\tabcolsep{2.25pt}
\caption{Reconstruction error.}
\label{tab:reconstruction_all}
\begin{tabular}{lcccc}
\toprule
\textbf{Method}                    & \textbf{dSprites}        & \textbf{3D Shapes}                        & \textbf{3D Faces}                        & \textbf{MPI 3D}                                                      \\ \midrule
 
VAE & $8.67 \pm 0.29$ & $3494.10 \pm 3.27$ & $1374.42 \pm 3.38$ & $3879.75 \pm 0.49$\\
VAE + TopDis (ours) & $9.54 \pm 0.19$ & $3489.53 \pm 1.50$ & $1376.22 \pm 0.32$ & $3879.89 \pm 0.51$ \\ \midrule

$\beta$-VAE & $12.97 \pm 0.50$ & $3500.60 \pm 13.59$ & $1379.64 \pm 0.19$ & $3888.84 \pm 2.45$ \\
$\beta$-VAE + TopDis (ours) & $13.75 \pm 0.63$ & $3495.76 \pm 6.54$ & $1380.10 \pm 0.19$ & $3886.57 \pm 0.81$\\ \midrule

FactorVAE & $14.65 \pm 0.41$ & $3501.53 \pm 13.43$ & $1488.26 \pm 4.47$ & $3884.31 \pm 0.59$ \\ 
FactorVAE + TopDis (ours) & $14.72 \pm 0.49$ & $3504.42 \pm 9.98$ & $1377.93 \pm 3.47$ & $3885.74 \pm 0.82$ \\ \midrule

$\beta$-TCVAE & $17.87 \pm 0.56 $ & $3492.25 \pm 5.79$ & $1375.03 \pm 3.41$ & $3891.03 \pm 1.41$\\ 
$\beta$-TCVAE + TopDis (ours) & $17.32 \pm 0.31 $ & $3495.13 \pm 2.49$ & $1376.21 \pm 3.09$ & $3889.34 \pm 1.97$\\ \midrule

ControlVAE & $15.32 \pm 0.47$ & $3499.61 \pm 12.13$ & $1404.42 \pm 5.01$ & $3889.81 \pm 0.43$ \\ 
ControlVAE + TopDis (ours) & $14.91 \pm 0.39$ & $3500.28 \pm 10.73$ & $1389.42 \pm 4.47$ & $3889.24 \pm 0.50$ \\ \midrule

DAVA & $36.41 \pm 2.03$ & $3532.56 \pm 14.14$ & $1403.77 \pm 0.99$ & $3890.42 \pm 2.15$ \\
DAVA + TopDis (ours) & $26.03 \pm 2.51$ & $3537.39 \pm 40.52$ & $1403.20 \pm 0.49$ & $3893.41 \pm 3.48$ \\
\bottomrule
\end{tabular}
\end{center}
\end{table}

\section{Training Statistics}

\begin{figure}[!ht]
\centering
\includegraphics[width=0.75\linewidth]{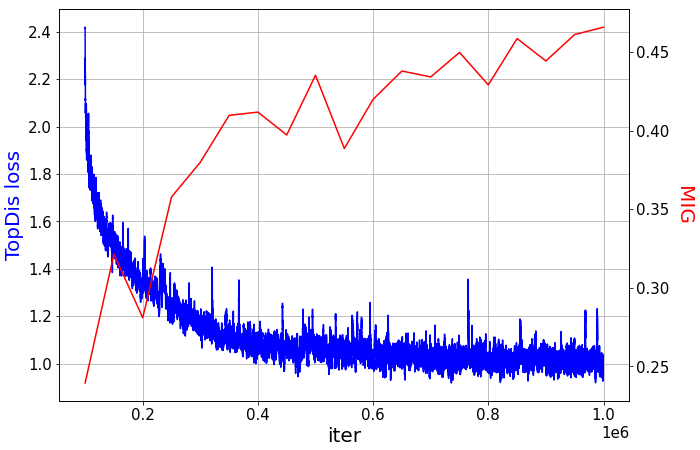}
\captionof{figure}{Training curves of TopDis loss and MIG for FactorVAE + TopDis on MPI 3D dataset.}
\label{fig:mpi3d_rtd_VS_MIG_1}
\end{figure}

Figure \ref{fig:mpi3d_rtd_VS_MIG_1} demonstrates that TopDis loss decreases during training  and has good negative correlation with MIG score, as expected. TopDis score was averaged with a sliding window of size $500$, MIG was calculated every $50 000$ iterations.

Also, we provide the TopDis scores for different axes in the latent space for the case of BetaVAE + TopDis and 3D Shapes dataset in Table \ref{tab:topdis_axes}. We randomly sample a batch with a fixed factor of variation. Following the definition of TopDis, we shift this batch with our shift in the probability space and compute RTD between the original and shifted batches. We average the result over different batch samples. As it can be seen, the values are in the same range for all factors, and none of the factors incurs much higher value of TopDis than others.

\begin{table}
\begin{center}
\setlength\tabcolsep{4pt}
\caption{TopDis scores for different axes in the latent space for BetaVAE + TopDis on 3D Shapes.}
\label{tab:topdis_axes}
\begin{tabular}{lcccccc}
\toprule
\textbf{Factor}                    & \textbf{Floor hue}        & \textbf{Wall hue}                        & \textbf{Object hue}                        & \textbf{Object size} & \textbf{Object shape} & \textbf{Orientation}       \\ \midrule
TopDis & $1.335 \pm 0.134$ & $0.926 \pm 0.086$ & $1.207 \pm 0.144$ &	$1.115 \pm 0.138$ &	$0.678 \pm 0.029$ & $0.763 \pm 0.067$ \\ \midrule
\end{tabular}
\end{center}
\end{table}

\section{More on Related Work}

In Table \ref{tbl:paper_baselines}, we compare our results with another recent state-of-the-art method, TCWAE
\citep{gaujac2021learning}.
Since there is no code available to replicate their results, we present the values from the original papers. The architecture and training setup were essentially identical to what is described in this paper.

In Table \ref{tbl:comparison_hfs}, we provide a comparison with HFS \cite{roth2022disentanglement}. As our TopDis, HFS loss is used in addition to standard disentanglement methods. Below we show the DCI disentanglement score for a subset of methods common in our paper and \cite{roth2022disentanglement}, numbers are taken from \cite{roth2022disentanglement}. TopDis always has a higher score. 

\begin{table}
\begin{center}
\setlength\tabcolsep{2.25pt}
\caption{Comparison with the TCWAE method}
\label{tbl:paper_baselines}
\begin{tabular}{lcccc}
\toprule
\textbf{Method}                    & \textbf{FactorVAE score}        & \textbf{MIG}                        & \textbf{SAP}                        & \textbf{DCI, dis.}       \\ \midrule
\multicolumn{5}{c}{dSprites}                                                                                                              \\ \midrule
TCWAE & $0.76 \pm 0.03$ & $0.32 \pm 0.04$ & $ 0.072 \pm 0.004$ & - \\
FactorVAE + TopDis (ours) & $\mathbf{0.82 \pm 0.04}$ & $\mathbf{0.36 \pm 0.03}$ & $\mathbf{0.082 \pm 0.001}$ & $\mathbf{0.52 \pm 0.04}$ \\ \midrule
\end{tabular}
\end{center}
\end{table}

\begin{table}[tbp]
\begin{center}
\caption{Comparison with the HFS loss by a DCI disentanglement score.}
\label{tbl:comparison_hfs}
\begin{tabular}{lccc}
\toprule
\textbf{Method} & \textbf{dSprites} & \textbf{Shapes3D} & \textbf{MPI 3D} \\
\midrule
$\beta$-VAE + HFS & 0.506 & 0.912 & 0.328\\ 
$\beta$-VAE + TopDis  & \textbf{0.556} & \textbf{0.998} & \textbf{0.337} \\ 
\midrule
$\beta$-TCVAE + HFS	& 0.499	& 0.857 & 0.328 \\
$\beta$-TCVAE + TopDis & \textbf{0.531} & \textbf{0.901} & \textbf{0.356} \\
\bottomrule
\end{tabular}
\end{center}
\end{table}

\section{Unsupervised Discovery of Disentangled Directions in StyleGAN}
\label{app:stylgan}
\begin{figure}[!ht]
\centering
\includegraphics[width=0.8\linewidth]{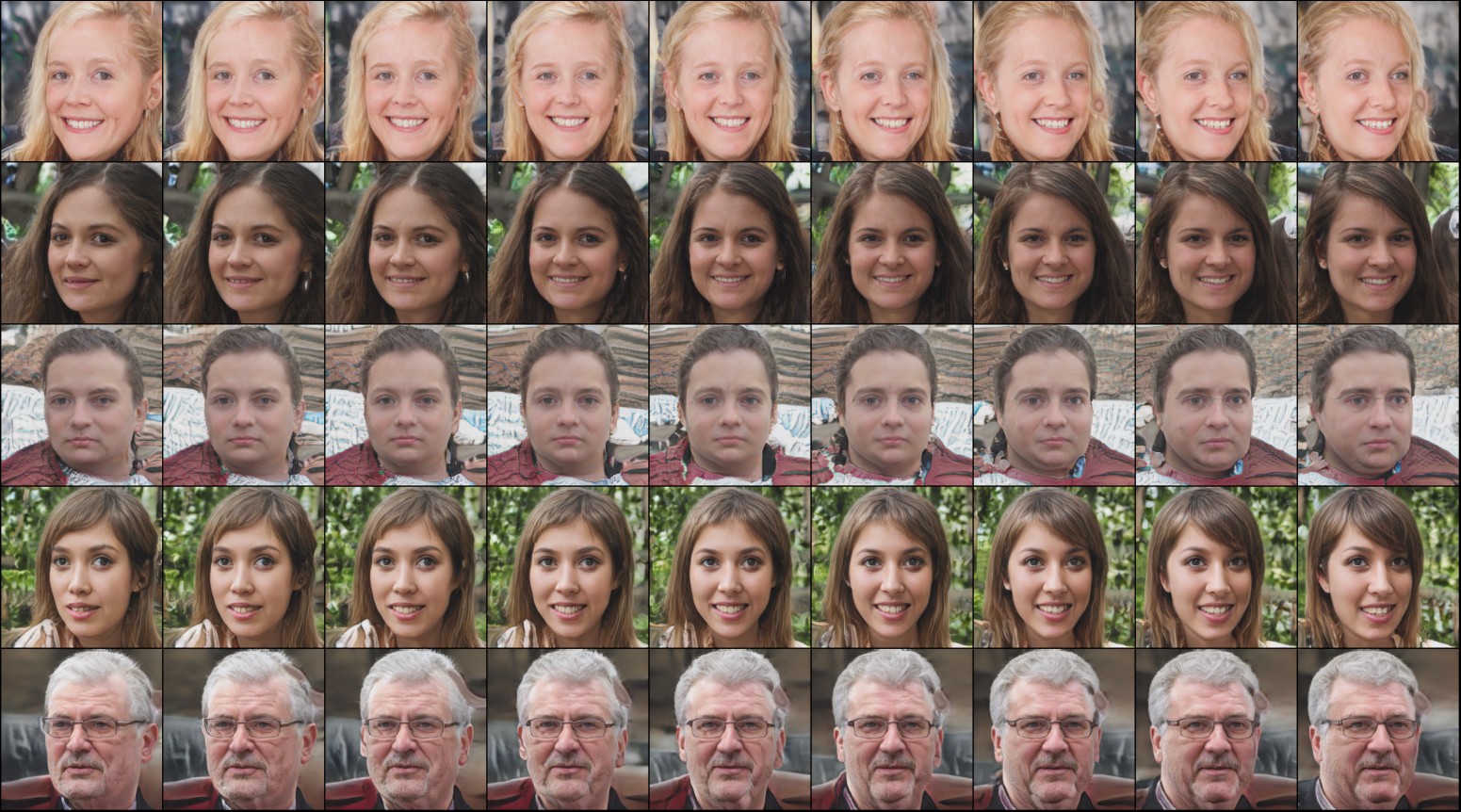}
\captionof{figure}{StyleGAN, change of an azimuth.}
\label{fig:stylegan_azimuth}
\end{figure}

\begin{figure}[!ht]
\centering
\includegraphics[width=0.8\linewidth]{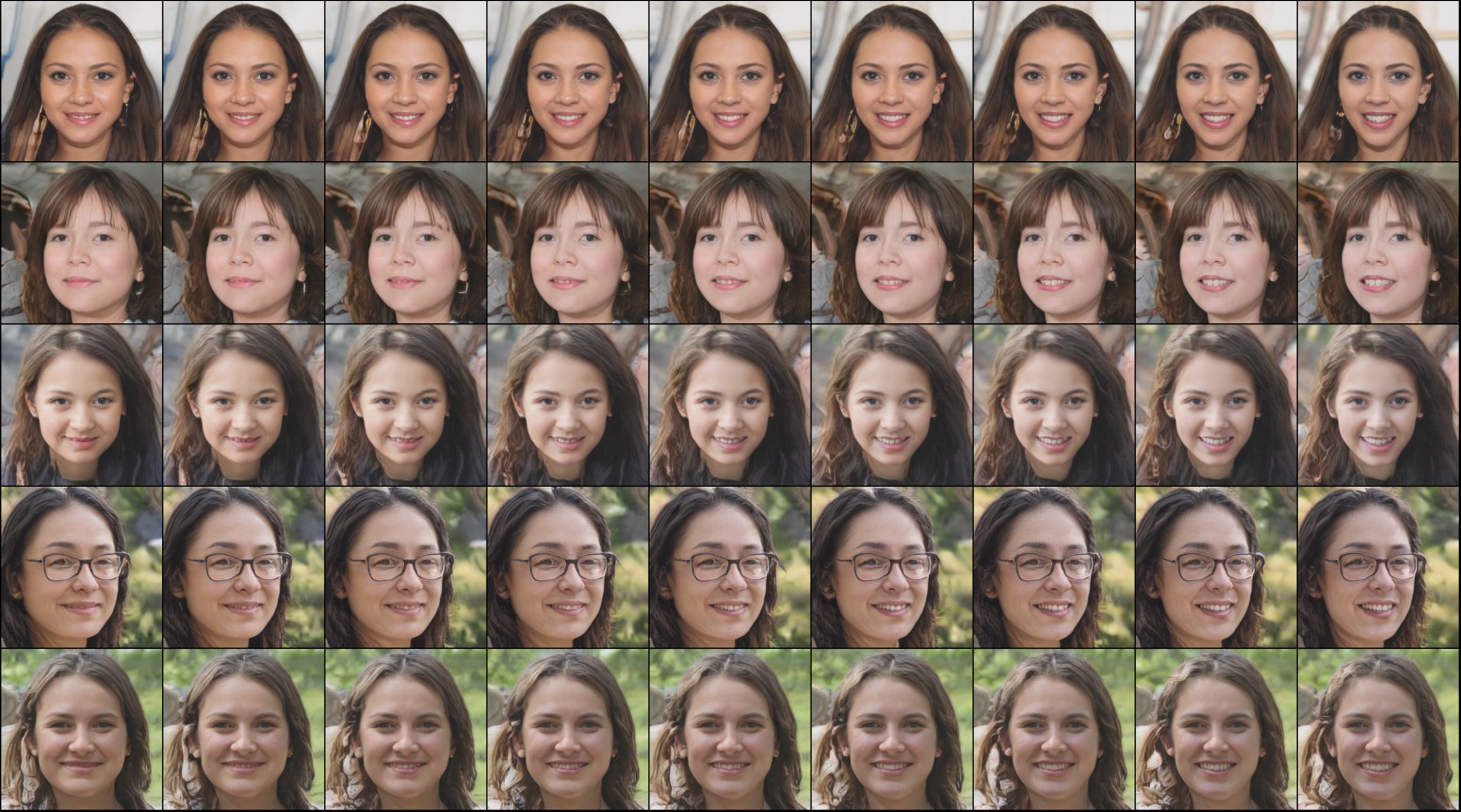}
\captionof{figure}{StyleGAN, change of a smile.}
\label{fig:stylegan_mouth}
\end{figure}

\begin{figure}[!ht]
\centering
\includegraphics[width=0.8\linewidth]{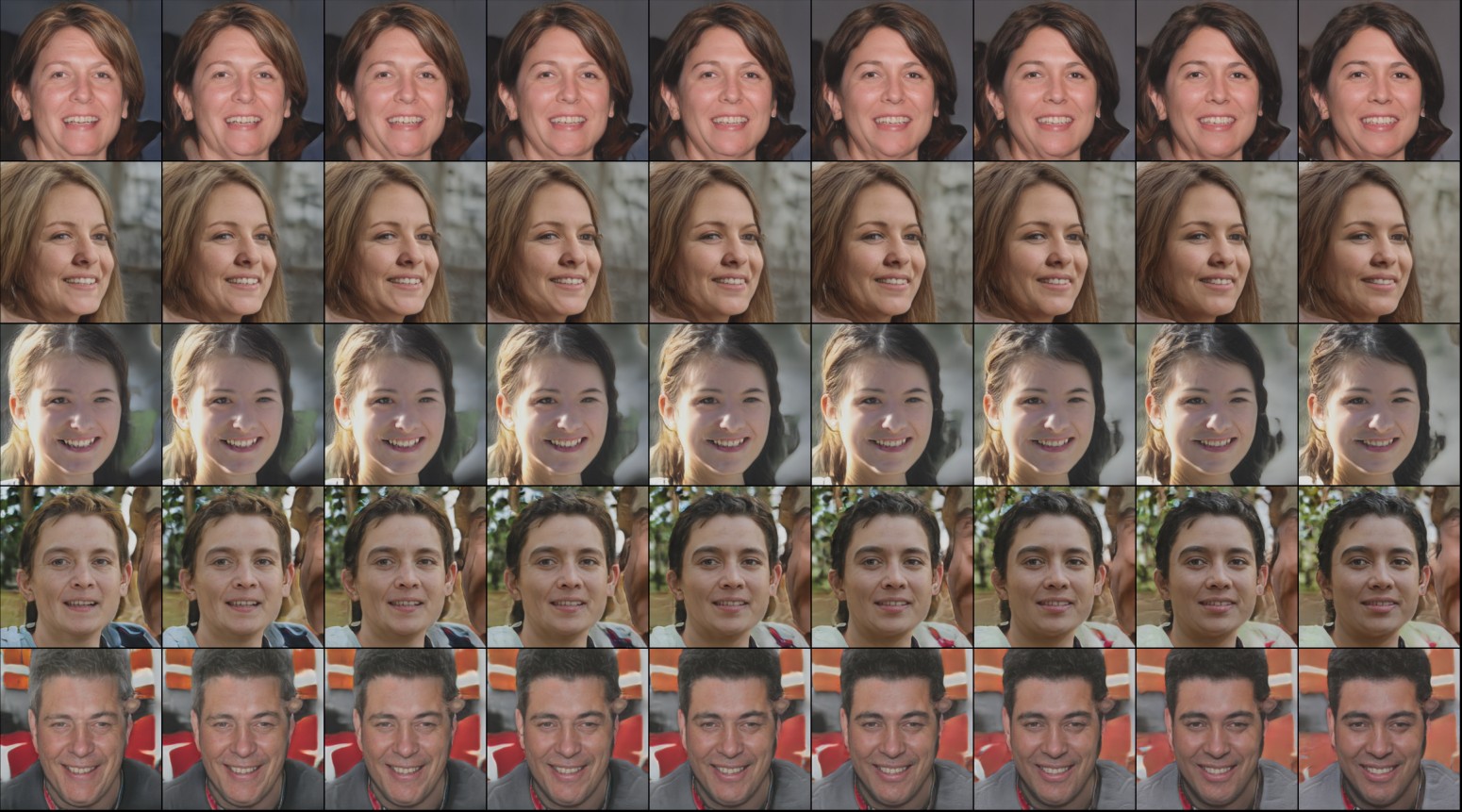}
\captionof{figure}{StyleGAN, change of a hair color.}
\label{fig:stylegan_hair}
\end{figure}

We perform additional experiments to study the ability of the proposed topology-based loss to infer disentangled directions in a pretrained GAN.
In experiments, we used StyleGAN \citep{karras2019style}\footnote{we used a PyTorch reimplementation from:\\ \url{https://github.com/rosinality/style-based-gan-pytorch}.}.
The unsupervised directions were explored in the style 
 space $\mathcal{Z}$. To filter out non-informative directions we followed the approach from \citet{harkonen2020ganspace} and selected top 32 directions by doing PCA for the large batch of data in the style space.
Then, we selected the new basis $n_i, \; i = 1, 
\ldots, 32$ in this subspace, starting from a random initialization. 
Directions $n_i$ were selected sequentially by minimization of RTD along shifts in $\mathcal{Z}$ space:
\begin{equation*}
 \operatorname{RTD}(Gen_k(Z), Gen_k(Z+c n_i)),   
\end{equation*}
where $Gen_k(\cdot)$ is the $k-$layer of the StyleGAN generator (we used $k=3$). After each iteration the Gram–Schmidt orthonormalization process for $n_i$ was performed.
We were able to discover at least 3 disentangled directions: azimuth (Fig. \ref{fig:stylegan_azimuth}), smile (Fig. \ref{fig:stylegan_mouth}), hair color (Fig. \ref{fig:stylegan_hair}).

\section{Extra Details on the Significance of TopDis Effect}
\label{sec:mann-witney_crit}
In order to accurately assess the impact of the TopDis term, we employed a consistent set of random initializations. This approach was adopted to eliminate potential confounding factors that may arise from disparate initial conditions. This allowed us to attribute any observed improvements in disentanglement quality specifically to the inclusion of the TopDis term in our model. In Table \ref{tbl:significance} we demonstrate the consistent improvement across multiple runs.

\begin{table}
\begin{center}
\setlength\tabcolsep{1pt}
\caption{Evaluation of the proposed FactorVAE + TopDis on the benchmark datasets for separate runs.}
\label{tbl:significance}
\begin{tabular}{lcccccccccccc}
\toprule
\textbf{Method}                    & \multicolumn{3}{c}\textbf{FactorVAE score}        & \multicolumn{3}{c}\textbf{MIG}                        & \multicolumn{3}{c}\textbf{SAP}                        & \multicolumn{3}{c}\textbf{DCI, dis.}       \\ \cmidrule(lr){2-4} \cmidrule(lr){5-7} \cmidrule(lr){8-10} \cmidrule(lr){11-13}
 & \text{run 1} & \text{run 2} & \text{run 3} & \text{run 1} & \text{run 2} & \text{run 3} & \text{run 1} & \text{run 2} & \text{run 3} & \text{run 1} & \text{run 2} & \text{run 3} \\ \midrule
\multicolumn{13}{c}{dSprites}                                                                                                              \\ \midrule
FactorVAE  & $\mathbf{0.856}$ & $0.830$ & $0.786$ & $\mathbf{0.341}$ & $0.308$ & $0.243$ & $0.054$ & $0.053$ & $0.051$ & $\mathbf{0.565}$ & $0.526$ & $\mathbf{0.509}$ \\ 
FactorVAE \\ + TopDis (ours) & $0.779$ & $\mathbf{0.845}$ & $\mathbf{0.847}$ & $0.331$ & $\mathbf{0.382}$ & $\mathbf{0.360}$ & $\mathbf{0.082}$ & $\mathbf{0.081}$ & $\mathbf{0.092}$ & $0.489$ & $\mathbf{0.571}$ & $0.503$ \\ \midrule
\multicolumn{13}{c}{3D Shapes}                                                                                                              \\ \midrule
FactorVAE & $0.901$ & $0.893$ & $\mathbf{1.000}$ & $0.678$ & $0.573$ & $\mathbf{0.867}$ & $0.055$ & $0.067$ & $0.175$ & $0.780$ & $0.772$ & $\mathbf{0.996}$ \\ 
FactorVAE \\ + TopDis (ours) & $\mathbf{0.924}$ & $\mathbf{1.000}$ & $\mathbf{1.000}$ & $\mathbf{0.810}$ & $\mathbf{0.787}$ & $0.739$ & $\mathbf{0.123}$ & $\mathbf{0.172}$ & $\mathbf{0.184}$ & $\mathbf{0.837}$ & $\mathbf{0.991}$ & $0.991$ \\ \midrule
\multicolumn{13}{c}{3D Faces}                                                                                                              \\ \midrule
FactorVAE & $\mathbf{1.000}$ & $\mathbf{1.000}$ & $\mathbf{1.000}$ & $0.597$ & $0.533$ & $0.649$ & $\mathbf{0.059}$ & $0.048$ & $0.076$ & $0.843$ & $0.840$ & $0.861$ \\ 
FactorVAE \\ + TopDis (ours) & $\mathbf{1.000}$ & $\mathbf{1.000}$ & $\mathbf{1.000}$ & $\mathbf{0.631}$ & $\mathbf{0.596}$ & $\mathbf{0.651}$ & $0.058$ & $\mathbf{0.051}$ & $\mathbf{0.077}$ & $\mathbf{0.859}$ & $\mathbf{0.835}$ & $\mathbf{0.907}$ \\ \midrule
\multicolumn{13}{c}{MPI 3D}                                                                                                              \\ \midrule
FactorVAE & $0.582$ & $0.651$ & $0.650$ & $0.323$ & $0.388$ & $0.341$ & $0.160$ & $0.230$ & $\mathbf{0.239}$ & $0.379$ & $0.435$ & $0.430$ \\ 
FactorVAE \\ + TopDis (ours) & $\mathbf{0.696}$ & $\mathbf{0.662}$ & $\mathbf{0.674}$ & $\mathbf{0.455}$ & $\mathbf{0.416}$ & $\mathbf{0.389}$ & $\mathbf{0.283}$ & $\mathbf{0.259}$ & $0.221$ & $\mathbf{0.505}$ & $\mathbf{0.464}$ & $\mathbf{0.437}$ \\
\bottomrule
\end{tabular}
\end{center}
\end{table}

\section{Training Details}
\label{app:hyperparams}
Following the previous work \citet{kim2018disentangling}, we used similar architectures for the encoder, decoder and discriminator, the same for all models. We set the latent space dimensionality to $10$. We normalized the data to $[0, 1]$ interval and trained $1$M iterations with batch size of $64$ and Adam \citep{kingma2015adam} optimizer. 
The learning rate for VAE updates was $10^{-4}$ for dSprites and MPI 3D datasets, $10^{-3}$ for 3D Shapes dataset, and $2\times10^{-4}$ for 3D faces and CelebA datasets, $\beta_1=0.9$, $\beta_2=0.999$, while the learning rate for discriminator updates was $10^{-4}$ for dSprites, 3D Faces, MPI 3D and CelebA datasets, $10^{-3}$ for 3D Shapes dataset, $\beta_1=0.5$, $\beta_2=0.9$ for discriminator updates.
In order to speed up convergence, we first trained the model without TopDis loss for a certain number of iterations and then continued training with TopDis loss.
We also fine-tuned the hyperparameter $\gamma$ over set commonly used in the literature \citep{kim2018disentangling, locatello2019challenging, ridgeway2018learning} to achieve the best performance on the baseline models.

The best performance found hyperparameters are the following: 
\begin{itemize}
\item dSprites. $\beta$-VAE: $\beta=2$, $\beta$-VAE + TopDis: $\beta=2, \gamma=4$, FactorVAE: $\gamma=20$, FactorVAE + TopDis: $\gamma_1=5, \gamma_2=5$, $\beta$-TCVAE: $\beta=6$, $\beta$-TCVAE+ TopDis: $\beta=6, \gamma=5$, DAVA + TopDis: $\gamma=5$;

\item 3D Shapes. $\beta$-VAE: $\beta=2$, $\beta$-VAE + TopDis: $\beta=2, \gamma=1$, FactorVAE: $\gamma=30$, FactorVAE + TopDis: $\gamma_1=5, \gamma_2=5$, $\beta$-TCVAE: $\beta=4$, $\beta$-TCVAE + TopDis: $\beta=4, \gamma=5$, DAVA + TopDis: $\gamma=3$;

\item 3D Faces. $\beta$-VAE: $\beta=2$, $\beta$-VAE + TopDis: $\beta=2, \gamma=1$, FactorVAE: $\gamma=5$, FactorVAE + TopDis: $\gamma_1=5, \gamma_2=5$, $\beta$-TCVAE: $\beta=6$, $\beta$-TCVAE + TopDis: $\beta=6, \gamma=5$, DAVA + TopDis: $\gamma=2$;

\item MPI 3D. $\beta$-VAE: $\beta=2$, $\beta$-VAE + TopDis: $\beta=2, \gamma=1$, FactorVAE: $\gamma=10$, FactorVAE + TopDis: $\gamma_1=5, \gamma_2=6$, $\beta$-TCVAE: $\beta=6$, $\beta$-TCVAE + TopDis: $\beta=6, \gamma=5$, DAVA + TopDis: $\gamma=5$;

\item CelebA. FactorVAE: $\gamma=5$, FactorVAE + TopDis: $\gamma_1=5, \gamma_2=2$;
\end{itemize}
For the ControlVAE and ControlVAE+TopDis experiments\footnote{ \url{https://github.com/shj1987/ControlVAE-ICML2020}.}, we utilized the same set of relevant hyperparameters as in the FactorVAE and FactorVAE+TopDis experiments. Additionally, ControlVAE requires an expected KL loss value as a hyperparameter, which was set to KL=18, as in the original paper. It should also be noted that the requirement of an expected KL loss value is counterintuitive for an unsupervised problem, as this value depends on the number of true factors of variation.
For the DAVA and DAVA + TopDis experiments\footnote{\url{https://github.com/besterma/dava}}, we used the original training procedure proposed in \citep{estermann2023dava}, adjusting the batch size to $64$ and number of iteration to $1$M to match our setup. Please, refer to our GitHub repository for further details. 

\section{Computational Complexity}
\label{app:complexity}
The complexity of the $\mathcal{L}_{TD}$ is formed by the calculation of RTD. For the batch size $N$, object dimensionality $C\times H \times W$ and latent dimensionality $d$, the complexity is $O(N^{2}(CHW+d))$, because all the pairwise distances in a batch should be calculated. 
The calculation of the RTD itself is often quite fast for batch sizes $\leq256$ since the boundary matrix is typically sparse for real datasets \citep{barannikov2021representation}.
Operations required to RTD differentiation do not take extra time. For RTD calculation and differentiation, we used  GPU-optimized software.

\section{Formal Definition of Representation Topology Divergence (RTD)}
\label{app:formal_rtd}

Data points in a high-dimensional space are often concentrated near a low-dimensional manifold  \citep{goodfellow2016deep}. The manifold's topological features can be represented via  Vietoris-Rips simplicial complex, a union of simplices whose vertices are points at a distance smaller than a threshold $\alpha$.

We define the weighted graph $\mathcal{G}$ with data points as vertices and the distances between data points $d({A_iA_j})$ as edge weights. The Vietoris-Rips complex at the threshold $\alpha$ is then:
$$
   \text{VR}_\alpha(\mathcal{G})=\left\{\{{A_{i_0}},\ldots,{A_{i_k}}\}, A_i \in \text{Vert}(\mathcal{G}) \; \vert \; d({A_iA_j}) \leq \alpha \right\},
$$

The vector space $C_k$ consists of all formal linear combinations of the $k$-dimensional simplices from $\text{VR}_{\alpha}(\mathcal{G})$ with modulo 2 arithmetic.
 The boundary operators $\partial_k: C_k \to C_{k-1}$ maps each simplex to the sum of its facets. The $k$-th homology group $H_k=ker(\partial_k)/im(\partial_{k+1})$ represents $k-$dimensional topological features.

Choosing $\alpha$ is challenging, so we analyze all $\alpha>0$. This creates a filtration of nested Vietoris-Rips complexes. We track the "birth" and "death" scales, $\alpha_{b},\alpha_{d}$, of each topological feature, defining its persistence as $\alpha_{d}-\alpha_{b}$. The sequence of the intervals $[\alpha_{b},\alpha_{d}]$ for basic features forms the persistence barcode \citep{Barannikov1994,chazal2017introduction}.

The standard persistence barcode analyzes a single point cloud $X$. The Representation Topology Divergence (RTD) \citep{barannikov2021representation} was introduced to measure the multi-scale topological dissimilarity between two point clouds $X, \tilde{X}$. This is done by constructing an auxilary graph $\hat{\mathcal{G}}^{w, \tilde{w}}$  whose Vietoris-Rips complex measures the difference between Vietoris-Rips complexes $\text{VR}_{\alpha}(\mathcal{G}^{w})$ and $\text{VR}_{\alpha}(\mathcal{G}^{\tilde{w}})$, where $w, \tilde{w}$ are the distance matrices of $X, \tilde{X}$.
The auxiliary graph  $\hat{\mathcal{G}}^{w, \tilde{w}}$ has the double set of vertices and the edge weights matrix 
$
\begin{pmatrix}
  0 & (w_{+})^\intercal \\\
 w_{+} & \min(w, \tilde{w}) 
 \end{pmatrix},
$
where $w_{+}$ is the $w$ matrix with lower-triangular part replaced by $+\infty$.

 The   \textit{R-Cross-Barcode$_k (X,\tilde{X})$} is the persistence barcode of the filtered simplicial complex 
 $\text{VR}(\hat{\mathcal{G}}^{w, \tilde{w}})$.
 $\text{RTD}_k(X, \tilde{X})$ equals the sum of intervals' lengths in \textit{R-Cross-Barcode}$_k (X, \tilde{X})$ and measures its closeness to an empty set, with longer lifespans indicating essential features.  $\text{RTD}(X,\tilde{X})$ is the half-sum  $\text{RTD}(X,\tilde{X}) = \nicefrac{1}{2}(\text{RTD}_1(X, \tilde{X}) + \text{RTD}_1(\tilde{X},X)).$

 \section{Symmetry Group(oid) Action}
\label{groupoid}
   A groupoid is a mathematical structure that generalizes the concept of a group. It consists of a set \( G \) along with a partially defined binary operation. Unlike groups, the binary operation in a groupoid is not required to be defined for all pairs of elements. More formally, a groupoid is a set \( G \) together with a binary operation \( \cdot : G \times G \rightarrow G \) that satisfies the following conditions for all \( a, b, c \) in \( G \) where the operations are defined:
   1) Associativity: \( (a \cdot b) \cdot c = a \cdot (b \cdot c) \);
   2) Identity:  there is an element \( e \) in \( G \) such that \( a \cdot e = e \cdot a = a \) for each \( a \) in \( G \);
   3) Inverses: for each \( a \) in \( G \), there is an element \( a^{-1} \) in \( G \) such that \( a \cdot a^{-1} = a^{-1} \cdot a = e \). 

A Lie groupoid is a groupoid that has additional structure of a manifold, together with smooth structure maps. These maps are required to satisfy certain properties analogous to those of a groupoid, but in a smooth category. See \citet{weinstein1996groupoids} for details.

\section{Gradient Orthogonalization Ablation Study}
\label{app:ablation}

Gradient orthogonalization is a technique to optimize a sum of two losses, which ensures that decreasing the second loss doesn’t conflict with the decrease of the first loss. Considering our TopDis loss term $L_{TD}$ and the reconstruction loss term $L_{rec}$, we take the projection of $\nabla L_{TD}$ on orthogonal space w.r.t the gradient $\nabla L_{rec}$ if their scalar product is negative. Moving within this direction allows the model parameters to get closer to the low error region for 
$L_{TD}$ while preserving the reconstruction quality at the same time.

\begin{figure}[ht!]
\begin{center}
\centering
\begin{subfigure}[t]{0.49\textwidth}
    \centering
    \includegraphics[width=0.8\textwidth]{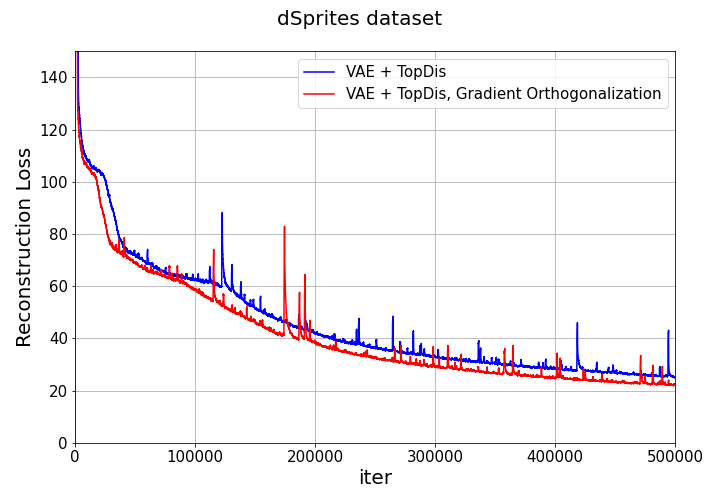}
    \label{fig:dsprites_ablation}
\end{subfigure}
\begin{subfigure}[t]{0.49\textwidth}
    \centering
    \includegraphics[width=0.8\textwidth]{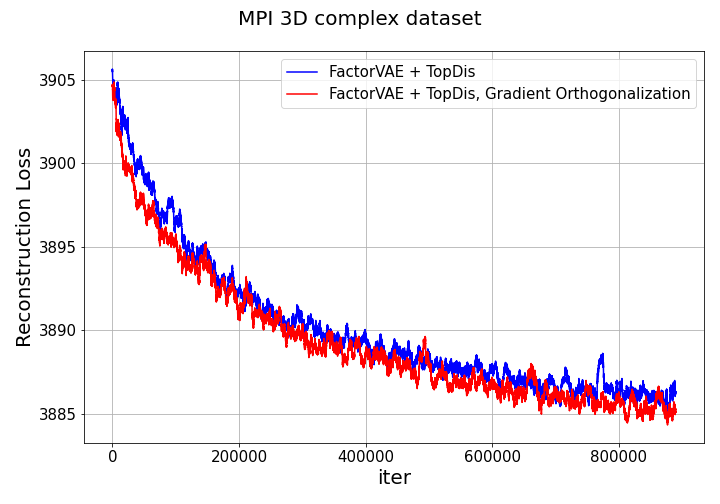}
    \label{fig:mpi3d_ablation}
\end{subfigure}%
\caption{Effect of gradient orthogonalization on reconstruction loss. Left: VAE+TopDis, dSprites. Right: FactorVAE+TopDis, MPI 3D.}
\label{fig:ort}
\end{center}
\end{figure}

We have performed the experiments concerning the ablation study of gradient orthogonalization technique. First, we evaluate the effect of gradient orthogonalization when integrating TopDis into the classical VAE model on dSprites, see Figure \ref{fig:ort} and Table \ref{tab:ablation}. We conduct this experiment to verify the gradient orthogonalization technique in the basic setup when additional terms promoting disentanglement are absent. Second, we evaluate the effect of gradient orthogonalization when integrating TopDis to FactorVAE on the MPI3D dataset. This experiment verifies how gradient orthogonalization works for more complex data in the case of a more complicated objective. We highlight that adding the gradient orthogonalization results in lower reconstruction loss throughout the training. In particular, this may be relevant when the reconstruction quality is of high importance. Similar technique was applied for continual and multi-task learning \citet{farajtabar2020orthogonal, suteu2019regularizing, yu2020gradient}.

\begin{table}[ht!]
\begin{center}
\setlength\tabcolsep{2.5pt}
\caption{Effect of gradient orthogonalization on disentanglement.}
\label{table2}
\begin{tabular}{lcccc}
\toprule
\textbf{Method}                    & \textbf{FactorVAE}        & \textbf{MIG}                        & \textbf{SAP}                        & \textbf{DCI, dis.}       \\ \midrule
\multicolumn{5}{c}{dSprites}                                                                                                              \\ \midrule
VAE + TopDis, no gradient orthogonalization & $0.736$ & $0.098$ & $0.041$ & $0.202$ \\
VAE + TopDis, gradient orthogonalization & $0.723$ & $0.121$ & $0.031$ & $0.229$ \\ \midrule

\multicolumn{5}{c}{MPI 3D}                                                                                                              \\ \midrule
FactorVAE + TopDis, no gradient orthogonalization & $0.696$ & $0.455$ & $0.283$ & $0.505$\\
FactorVAE + TopDis, gradient orthogonalization & $0.707$ & $0.466$ & $0.288$ & $0.508$ \\ 
\bottomrule
\label{tab:ablation}
\end{tabular}
\end{center}
\end{table}

\section{Sensitivity Analysis}
\label{app:sensitivity}

The proposed TopDis approach has two hyperparameters - the weight of loss term $\gamma$ (see equation \ref{eq:objective} for details) and the value of the shift $C$ (see equation \ref{shifts} for details).
We provide the sensitivity analysis w.r.t. $\gamma$ for FactorVAE+TopDis on MPI3D-Real ($3\cdot 10^5$ training iterations), please see Table \ref{tab:sensitivity}. In Table {\ref{tab:sensitivity}}, $\gamma_{TD}$ denotes the weight $\gamma$ for the TopDis loss from equation {\ref{eq:objective}} while $\gamma_{TC}$ denotes the weight for the Total Correlation loss from the FactorVAE model (see \mbox{\cite{kim2018disentangling}} for details). In particular, $\gamma_{TC}=5, \gamma_{TD}=0$ corresponds to plain FactorVAE model. In practice, when integrating the TopDis loss, we first search for the set of the hyperparameters for the base model (VAE, FactorVAE, ControlVAE, etc.), and then tune the weight for the TopDis loss only. This strategy demonstrates the outperforming results in most cases.

\begin{table}[h!]
\begin{center}
\setlength\tabcolsep{3pt}
\caption{Sensitivity analysis. $\gamma_{TD}$ denotes the weight for the TopDis loss (see equation \ref{eq:objective} for details) while $\gamma_{TC}$ denotes the weight for the Total Correlation loss from the FactorVAE model (see \cite{kim2018disentangling} for details).}
\label{tab:sensitivity}
\begin{tabular}{lccccc}
\toprule
\textbf{Method}                    & \textbf{FactorVAE}        & \textbf{MIG}                        & \textbf{SAP}                        & \textbf{DCI, dis.} & \textbf{Reconstruction}       \\ \midrule

\multicolumn{5}{c}{FactorVAE + TopDis, MPI 3D}                                                                                                              \\ \midrule

$\gamma_{TC}=5, \gamma_{TD}=0$ & $0.586 \pm 0.038$ & $0.300 \pm 0.020$ & $0.184 \pm 0.028$ & $0.357 \pm 0.001$ & $3888.96 \pm 0.94$\\
$\gamma_{TC}=5, \gamma_{TD}=3$ & $0.607 \pm 0.047$ & $0.320 \pm 0.003$ & $0.193 \pm 0.015$ & $0.401 \pm 0.025 $ & $3891.28 \pm 0.91$\\
$\gamma_{TC}=5, \gamma_{TD}=5$ & $0.605 \pm 0.048$ & $0.332 \pm 0.033$ & $0.205 \pm 0.025$ & $0.397 \pm 0.038$ & $3892.51 \pm 1.19$\\
$\gamma_{TC}=5, \gamma_{TD}=6$ & $0.605 \pm 0.051$ & $0.340 \pm 0.035$ & $0.207 \pm 0.036 $ & $0.412 \pm 0.026$ & $3892.17 \pm 0.54$\\
$\gamma_{TC}=5, \gamma_{TD}=7$ & $0.594 \pm 0.041$ & $0.297 \pm 0.042$ & $0.183 \pm 0.029$ & $0.362 \pm 0.050 $ & $3892.98 \pm 0.70$ \\ 
\bottomrule
\end{tabular}
\end{center}
\end{table}

Further, Table \ref{tab:sensitivity_C} provides analysis of performance for different values of $C$ from equation \ref{shifts} for FactorVAE+TopDis model on dSprites dataset (1M training iterations). In practice, we choose the value of $C$ to be the same across all datasets, and we found the choice $C=1/8$ demonstrates the best performance for all our experiments.

\begin{table}[h!]
\begin{center}
\setlength\tabcolsep{3pt}
\caption{Sensitivity analysis. $C$ denotes the value of shift for the TopDis loss (see equation \ref{shifts} for details).}
\label{tab:sensitivity_C}
\begin{tabular}{lccccc}
\toprule
\textbf{Method}                    & \textbf{FactorVAE}        & \textbf{MIG}                        & \textbf{SAP}                        & \textbf{DCI, dis.} & \textbf{Reconstruction}       \\ \midrule

\multicolumn{6}{c}{FactorVAE + TopDis, dSprites}                                                                                                              \\ \midrule

FactorVAE & $0.819 \pm 0.028$ & $0.295 \pm 0.049$ & $0.053 \pm 0.006$ & $0.534 \pm 0.029$ & $14.65 \pm 0.41$ \\
FactorVAE + TopDis, $C = 1/16$ & $0.779 \pm 0.021$ & $0.344 \pm 0.029$ & $0.058 \pm 0.004$ & $0.528 \pm 0.032$ & $14.71 \pm 0.47$\\
FactorVAE + TopDis, $C = 1/8$ &	$0.824 \pm 0.038$ &	$0.356 \pm 0.025$ &	$0.082 \pm 0.001$ &	$0.521 \pm 0.044$ & $14.72 \pm 0.49$\\
FactorVAE + TopDis, $C = 1/4$ &	$0.820 \pm 0.041$ & $0.340 \pm 0.033$ &	$0.058 \pm 0.002$ &	$0.525 \pm 0.062$ & $14.85 \pm 0.51$\\
\bottomrule
\end{tabular}
\end{center}
\end{table}

\section{Group(oid) Action Versus Constant Shift Ablation}
\label{app:shift_ablation}

To demonstrate the relevance of the proposed shift in latent codes (see Section \ref{section:shift} for details), we perform the ablation experiment for the FactorVAE + TopDis and the MPI 3D dataset. In this ablation, we replace the proposed shift (i.e. equation \ref{shifts}) in the latent space with the shift which is the same for all objects in the batch. To keep the reasonable magnitude of this shift, we take the shift to be proportional to the standard deviation of the batch in a chosen latent dimension. Table \ref{tab:abl_shift} reveals that although the constant shift has positive effect on model's performance in comparison with FactorVAE, it results in worse performance than FactorVAE + TopDis with the proposed shift. This example illustrates the effectiveness of the proposed shift procedure.

\begin{table}[h!]
\begin{center}
\setlength\tabcolsep{3pt}
\caption{Ablation experiment for the proposed shift in probability space (see equation \ref{shifts} for details). \textit{Const shift} denotes the shift of a reasonable magnitude that is the same for all the objects in the batch.}
\label{tab:abl_shift}
\begin{tabular}{lcccc}
\toprule
\textbf{Method}                    & \textbf{FactorVAE}        & \textbf{MIG}                        & \textbf{SAP}                        & \textbf{DCI, dis.}       \\ \midrule

\multicolumn{5}{c}{FactorVAE + TopDis, MPI 3D} \\ \midrule
FactorVAE &	$0.589 \pm 0.053$ &	$0.336 \pm 0.056$ & $0.179 \pm 0.052$ & $0.391 \pm 0.056$ \\
FactorVAE + TopDis & $0.665 \pm 0.041$ & $0.377 \pm 0.053$ & $0.238 \pm 0.040$ & $0.438 \pm 0.065$ \\
FactorVAE + TopDis const shift & $0.628 \pm 0.042$ & $0.342 \pm 0.059$ & $0.209 \pm 0.033$ & $0.418 \pm 0.038$ \\
\bottomrule
\end{tabular}
\end{center}
\end{table}

\section{RTD Differentiation}
\label{app:rtd_diff}

Here we gather details on RTD differentiation in order to use RTD as a loss in neural networks. 

Define $\Sigma$ as the set of all simplices in the filtration of the graph $VR(\hat{\mathcal{G}}^{w, \tilde{w}})$, and $\mathcal{T}_k$ as the set of all segments in $\textit{R-Cross-Barcode}_k(X, \hat{X})$. Fix (an arbitrary) strict order on $\mathcal{T}_k$.

There exists a function $ f_k: ~ \lbrace b_i, d_i \rbrace_{(b_i, d_i) \in \mathcal{T}_k} \rightarrow \Sigma$ that maps $b_i$ (or $d_i$) to simplices $\sigma$ (or $\tau$) whose addition leads to ``birth'' (or ``death'') of the corresponding homological class.

Thus, we may obtain the following equation for subgradient
$$
\frac{\partial ~ \text{RTD}(X, \hat{X})}{\partial \sigma} = \sum_{i \in \mathcal{T}_k} \frac{\partial \text{RTD}(X, \hat{X})}{\partial b_i}\mathbb{I}\lbrace f_k(b_i) = \sigma \rbrace + \sum_{i \in \mathcal{T}_k} \frac{\partial \text{RTD}(X, \hat{X})}{\partial d_i}\mathbb{I}\lbrace f_k(d_i) = \sigma \rbrace
$$

Here, for any $\sigma$ no more than one term has non-zero indicator.

$b_i$ and $d_i$ are just the filtration values at which simplices $f_k(b_i)$ and $f_k(d_i)$ join the filtration. They depend on weights of graph edges as
$$
g_k(\sigma) = \max_{i, j \in \sigma} m_{i, j}
$$

This function is differentiable \citep{leygonie2021framework} 
and so is $f_k \circ g_k$. Thus we obtain the subgradient:
$$
\frac{\partial ~ \text{RTD}(X, \hat{X})}{\partial m_{i, j}} = \sum_{\sigma \in \Sigma}\frac{\partial ~ \text{RTD}(X, \hat{X})}{\partial \sigma}  \frac{\partial \sigma}{\partial m_{i, j}}.
$$

The only thing that is left is to obtain subgradients of $\text{RTD}(X, \hat{X})$ by points from $X$ and $\hat{X}$ .
Consider (an arbitrary) element $m_{i, j}$ of matrix $m$. There are 4 possible scenarios:
\begin{enumerate}
    \item $i, j \leq N$, in other words $m_{i, j}$ is from the upper-left quadrant of $m$. Its length is constant and thus $\forall l: \frac{\partial m_{i, j}}{\partial X_l} =  \frac{\partial m_{i, j}}{\partial \hat{X_l}} = 0$.
    \item $i \leq N < j$, in other words $m_{i, j}$ is from the upper-right quadrant of $m$. Its length is computed as Euclidean distance and thus $\frac{\partial m_{i, j}}{\partial X_i} = \frac{X_i - X_{j - N}}{\|X_i - X_{j - N}\|_2}$ (similar for $X_{N - j}$).
    \item $j \leq N < i$, similar to the previous case.
    \item $N < i, j $, in other words $m_{i, j}$ is from the bottom-right quadrant of $m$. Here we have subgradients like
    $$\frac{\partial m_{i, j}}{\partial X_{i - N}} = \frac{X_{i - N} - X_{j - N}}{\|X_{i - N} - X_{j - N}\|_2}\mathbb{I}\lbrace w_{i - N, j - N} < \hat{w}_{i - N, j - N}\rbrace $$
    Similar for $X_{j-N}, \hat{X}_{i-N}$ and $\hat{X}_{j-N}$.
\end{enumerate}

Subgradients $\frac{\partial ~ \text{RTD}(X, \hat{X})}{\partial X_{i}}$ and $\frac{\partial ~ \text{RTD}(X, \hat{X})}{\partial \hat{X}_{i}}$ can be derived from the before mentioned using the chain rule and the formula of full (sub)gradient. Now we are able to minimize $\text{RTD}(X, \hat{X})$ by methods of (sub)gradient optimization. 

\section{Discussing the Definition of Disentangled Representation.}

Let $X\subset \mathbb{R}^{N_x\times N_y}$ denotes the dataset consisting of  $N_x\times N_y$ pixels pictures containing a disk of various color with fixed disk radius $r$ and the center of the disks situated at an arbitrary point $x,y$. Denote $\rho_X$ the uniform distribution over the coordinates of centers of the disks and the colors. 
Let $G_x\times G_y\times G_c$ be the commutative group of symmetries of this data distribution,  $G_x\times G_y$ is the position change acting (locally) via
\begin{equation*}
(a,b):(x,y,c)\mapsto (x+a,y+b,c)  
\end{equation*} and $G_z$ is changing the colour along the colour circle $\theta:(x,y,c)\mapsto (x,y,c+\theta\mod 2\pi)$. 
Contrary to \citet{higgins2018towards}, section 3,  we do not assume the 
gluing of the opposite sides of our pictures, which is closer to real world situations. 
Notice that, as a consequence of this,  each group element from $G_x\times G_y$ can act only on a subset of $X$, so that the result is still situated inside $N_x\times N_y$ pixels picture. This mathematical structure when each group element has its own set of points on which it acts, is called groupoid, we discuss this notion in more details in Appendix \ref{groupoid}. 

The outcome of  disentangled learning in such case are the encoder  $h:X\to Z$  and the decoder $f:Z\to X$ maps with $Z=\mathbb{R}^3$, $f\circ h=\text{Id}$, together with symmetry group(oid) $G$ actions on $X$ and $Z$, such that a) the encoder-decoder maps preserve the distributions, which are the distribution $\rho_X$ describing the dataset $X$ and the standard in VAE learning $N(0,1)$ distribution in latent space $Z$;
b) the decoder and the encoder maps are equivariant with respect to the symmetry group(oid) action, where the action on the latent space is defined as shifts of latent variables; the group action preserves the dataset distribution $X$ therefore the group(oid) action shifts on the latent space must  preserve the standard $N(0,1)$ distribution on latent coordinates, i.e. they must act via the formula \ref{shifts}.

\textbf{Connection with disentangled representations in which the symmetry group latent space action is \emph{linear}.}
The normal distribution arises naturally as the projection to an axis of the uniform distribution on a very high dimensional sphere $S^N\subset \mathbb{R}^{N+1}$.  Let a general symmetry compact Lie group $\hat{G}$ acts linearly on $\mathbb{R}^{N+1}$ and preserves the sphere $S^N$. Let $G^{ab}$ be a maximal commutative subgroup in $G$. Then the ambient space $\mathbb{R}^{N+1}$ decomposes into direct sum of subspaces $\mathbb{R}^{N+1}=\oplus_{\alpha}Z_{\alpha}$, on which $G^{ab}=\Pi_i G_i$,  acts  via rotations in two-dimensional space, and the orbit of this action is a circle $S^1\subset S^{N}$. If one chooses an axis in each such two-dimensional space then the projection to this axis gives a coordinate on the sphere $S^{N}$. And the group action of $G^{ab}$ decomposes into independent actions along these axes. In such a way, the disentangled representation in the  sense of Section \ref{TopGroup} can be obtained from the data representation with uniform distribution on the sphere/disk on which the symmetry group action is linear, and vice versa.

\section{Experiments with Correlated Factors}
\label{app:correlated_factors}

\begin{table}
\begin{center}
\setlength\tabcolsep{2.25pt}
\caption{Evaluation on the benchmark datasets with correlated factors}
\label{tbl:corr_factors_results}
\begin{tabular}{lcccc}
\toprule
\textbf{Method}                    & \textbf{FactorVAE score}        & \textbf{MIG}                        & \textbf{SAP}                        & \textbf{DCI, dis.}       \\ \midrule
\multicolumn{5}{c}{dSprites}                                                                                                              \\ \midrule
FactorVAE & $0.803 \pm 0.055$ & $0.086 \pm 0.026$ & $0.030 \pm 0.010$ & $0.216 \pm 0.044$ \\
FactorVAE + TopDis (ours) & \textbf{0.840 $\pm$ 0.011} & \textbf{0.103 $\pm$ 0.019} & \textbf{0.044 $\pm$ 0.014} & \textbf{0.270 $\pm$ 0.002} \\
\midrule
\multicolumn{5}{c}{3D Shapes}                                                                                                            \\ \midrule
FactorVAE & 0.949 $\pm$ 0.67 & 0.363 $\pm$ 0.100 & 0.083 $\pm$ 0.004 & 0.477 $\pm$ 0.116 \\
FactorVAE + TopDis (ours) &  \textbf{0.998 $\pm$ 0.001} & \textbf{0.403 $\pm$ 0.091} & \textbf{0.112 $\pm$ 0.013} & \textbf{0.623 $\pm$ 0.026}\\

\bottomrule
\end{tabular}
\end{center}
\end{table}

Table \ref{tbl:corr_factors_results} shows experimental results for disentanglement learning with confounders - one factor correlated with all others. The addition of the TopDis loss results in a consistent improvement of all quality measures. For experiments, we used the implementation of the ``shared confounders'' distribution from \cite{roth2022disentanglement}\footnote{\url{https://github.com/facebookresearch/disentangling-correlated-factors}} and the same hyperparameters as for the rest of experiments.

\section{Experiments with VAE + TopDis}
\label{app:vae_topdis}
In order to verify the proposed TopDis as a self-sufficient loss contributing to disentanglement, we add TopDis to the classical VAE \cite{kingma2013auto} objective as an additional loss term. As demonstrated by quantitative evaluation on the benchmark datasets in Table \ref{tbl:vae_topdis_results}, the addition of TopDis loss improves the quality of disentanglement as measured by FactorVAE score, MIG, SAP, DCI: on dSprites up to $+6\%$, $+17\%$, $+14\%$, $+25\%$, on 3D Shapes up to $+14\%$, $+35\%$, $+2\%$, on 3D Faces up to $+4\%$, $+2\%$, $+6\%$, $+2\%$, on MPI 3D up to $+7\%$, $+27\%$, $+37\%$, $+17\%$. For all the datasets and metrics, VAE+TopDis outperforms the classical VAE model. Besides, VAE+TopDis preserves the reconstruction quality as revealed by Table \ref{tab:reconstruction_all}.

\begin{table}
\begin{center}
\setlength\tabcolsep{2.25pt}
\caption{Evaluation on the benchmark datasets for VAE + TopDis}
\label{tbl:vae_topdis_results}
\begin{tabular}{lcccc}
\toprule
\textbf{Method}                    & \textbf{FactorVAE score}        & \textbf{MIG}                        & \textbf{SAP}                        & \textbf{DCI, dis.}       \\ \midrule
\multicolumn{5}{c}{dSprites}                                                                                                              \\ \midrule
VAE & $0.781 \pm 0.016$ & $0.170 \pm 0.072$ & $0.057 \pm 0.039$ & $0.314 \pm 0.072$ \\
VAE + TopDis (ours) & $\mathbf{0.833 \pm 0.068}$ & $\mathbf{0.200 \pm 0.119}$ & $\mathbf{0.065 \pm 0.009}$ & $\mathbf{0.394 \pm 0.132}$ \\
\midrule
\multicolumn{5}{c}{3D Shapes}                                                                                                            \\ \midrule
VAE & $\mathbf{1.0 \pm 0.0}$ & $0.729 \pm 0.070$ & $0.160 \pm 0.050$ & $0.952 \pm 0.023$ \\ 
VAE + TopDis (ours) & $\mathbf{1.0 \pm 0.0}$ & $\mathbf{0.835 \pm 0.012}$ & $\mathbf{0.216 \pm 0.020}$ & $\mathbf{0.977 \pm 0.023}$ \\
\midrule
\multicolumn{5}{c}{3D Faces}                                                                                                            \\ \midrule
VAE & $0.96 \pm 0.03$ & $0.525 \pm 0.051$ & $0.059 \pm 0.013 $ & $0.813 \pm 0.063$ \\ 
VAE + TopDis (ours) & $\mathbf{1.0 \pm 0.0}$ & $\mathbf{0.539 \pm 0.037}$ & $\mathbf{0.063 \pm 0.011}$ & $\mathbf{0.831 \pm 0.023}$ \\
\midrule
\multicolumn{5}{c}{MPI 3D}                                                                                                            \\ \midrule
VAE & $0.556 \pm 0.081$ & $0.280 \pm 0.059$ & $0.167 \pm 0.064$ & $0.346 \pm 0.029$ \\ 
VAE + TopDis (ours) & $\mathbf{0.595 \pm 0.055}$ & $\mathbf{0.358 \pm 0.022}$ & $\mathbf{0.229 \pm 0.022}$ & $\mathbf{0.407 \pm  0.025}$ \\

\bottomrule
\end{tabular}
\end{center}
\end{table}

\section{Motivation for Topological Feature Distance in Learning Disentangled Representations}\label{app:motivation}

To connect topological features with disentangled representations, the continuity and invertibility of the Lie transformations on the model distribution support
 are the key properties. Such transformations, known as homeomorphisms,  induce homology isomorphisms and hence preserve topological features. Minimizing TopDis ensures maximal preservation of these features by the Lie action. The following proposition further strengthens this relationship.

Let $W(RLT(d, k), RLT(d, k^\prime))$ denotes the Wasserstein distance between the RLT's of two data submanifolds conditioned on two values $k$ and $k^\prime$ of a generative factor $z_d$ \citep{zhou2020evaluating}. The proposition below proves that minimizing TopDis ensures a small topological distance between conditioned submanifolds, which is a requirement for disentanglement according to \citep{zhou2020evaluating}. 

\begin{proposition}\label{prop:wass_dist}
If $TopDis_i(z, a) < \epsilon$, for $a \leq \frac{1}{8}$, $i = 0, 1$, where $TopDis_i(z, a)$ is the $TopDis_i$ loss with the shift $z \to z^\prime$ along a generative factor $z_d$ with parameter $C = a$, then the Wasserstein distance between the RLT's for two data submanifolds conditioned on two values $k, k^\prime$ of $z_d$ satisfies $W(RLT(d, k), RLT(d, k^\prime)) < \frac{16}{\alpha_{max}}\epsilon$, where $\alpha_{max}$ is the constant from the definition of RLT.
\end{proposition}

\textit{Proof.}
Given a sample $z$ from the data submanifold with the generative factor $z_d$ conditioned on $z_d = k$, a shift along $z_d$ with some $C^\prime$ produces from $z$ a sample with the generative factor $z_d$ conditioned on $z_d = k^\prime$. For a small shift with $C^\prime \leq \frac{1}{8}$, the Wasserstein distance between RLT's of $z$ and $z^\prime$ is bounded by $\frac{1}{\alpha_{max}}(TopDis_0(z, C^\prime) + TopDis_1(z, C^\prime)) < \frac{2}{\alpha_{max}}\epsilon$. An arbitrary shift can be decomposed into no more than $8$ smaller shifts with $C^\prime \leq \frac{1}{8}$, and then the Wasserstein distance between RLT's of $z$ and $z^\prime$ is bounded similarly by $\frac{16}{\alpha_{max}}\epsilon$.

\section{Comparison with Previous Works on RTD}

The Representation Topology Divergence (RTD) was introduced in the paper \citep{barannikov2021representation}, as a tool for comparing two neural representations of the same set of objects.
An application of RTD to evaluation of interpretable directions in a simple synthetic dataset is
described briefly in loc.cit. The experiment in loc.cit. involves comparing the topological dissimilarity in data submanifolds corresponding to slices in the latent space. While in the current paper, we use axis-aligned traversals and samples from the whole data manifold. One of the crucial difference with loc.cit. 
is that, in the current paper, we are able to apply our topological loss TopDis (equation \ref{eq:td_def}) directly on the outcome of the decoder during the VAE optimization. This permits to propagate the gradients from the topological loss to the variational autoencoder weights. In \citet{barannikov2021representation}, RTD was used as a metric between two submanifolds conditioned on two values of a generative factor, in a static situation of an already disentangled model. During learning the outcome of the decoder never lies on such conditioned submanifold. This has led us further to the Lie group(oid) action and the Proposition \ref{prop:qzi} establishing equivalence of two approaches to disentanglement, via factorized distributions and via the Lie symmetry actions. We show in addition how to preserve the reconstruction quality via gradient orthogonalization. The effectiveness of gradient orthogonalization to improve the reconstruction (see Figure \ref{fig:ort}) and the disentanglement (Table \ref{tab:ablation}) is validated via ablation study. In the paper \citet{trofimov2023learning}, the RTD is used as an additional loss for dimensionality reduction, in order to preserve topology of original data in a latent space.

\section{A Remark on Inductive Bias}

In \citet{locatello2020asober}, the authors have shown that unsupervised disentanglement learning without any inductive bias is impossible. Next, we explain in what sense the proposed TopDis approach can be seen as an inductive bias. 

Each factor in a disentangled model has a symmetry Lie action that fixes other factors, and therefore each factor encodes variations that must cause only topologically mild changes in the pixel space. It's important to keep in mind that 'topologically' refers here to the topology of the clouds of data points in pixel space, rather than the topology of an object depicted on a specific dataset image. Regarding a categorical factor whose change produces objects in the pictures with objects' topology that might be dramatically different, such factors are expected to correspond to discrete symmetries on data manifolds. However, in our approach, these discrete symmetries are included into continuous families that correspond to continuous interpolation between the objects categories. These symmetry actions guarantee that such interpolations are the same for objects, for example, positioned at varying locations within an image. So a useful inductive bias is the fact that all disentangled factors cause continuous changes. For categorical factors, which might inherently possess only discrete symmetries, these discrete symmetries can be integrated into continuous interpolating families of symmetries. It's possible that these factors may exhibit slightly larger TopDis because data points corresponding to interpolations between dramatically different objects are absent from the dataset, potentially requiring more iterations to learn interpolations between object categories in a symmetrically uniform way.

\section{Visualization of Latent Traversals}
\label{app:traversals}

\begin{figure}[!ht]
\centering
\begin{subfigure}{0.49\textwidth}
  \centering
  \includegraphics[width=1\linewidth]{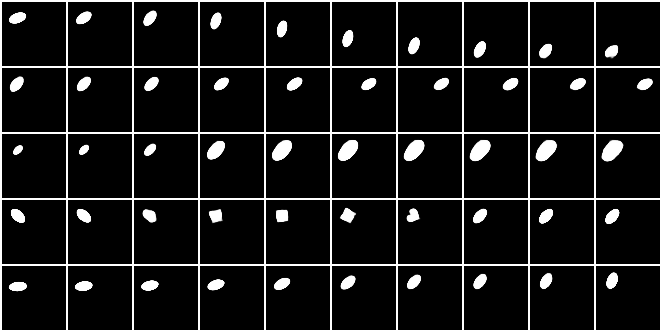}
    \caption{FactorVAE, dSprites.}
    \label{fig:FactorVAE_dsprites_traverse}
\end{subfigure}
\hfill
\begin{subfigure}{0.49\textwidth}
  \centering
  \includegraphics[width=1\linewidth]{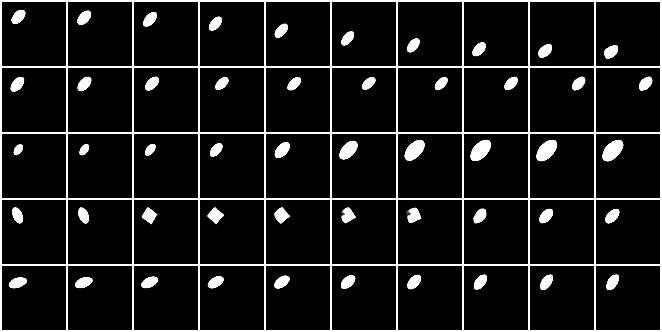}
    \caption{FactorVAE + TopDis, dSprites.}
    \label{fig:FactorVAE_RTD_dsprites_traverse}
\end{subfigure}

\begin{subfigure}{0.49\textwidth}
  \centering
  \includegraphics[width=1\linewidth]{pics/3dshapes_factorvae_nontrivial.png}
    \caption{FactorVAE, \mbox{3D Shapes}.}
    \label{fig:FactorVAE_3dshapes_traverse}
\end{subfigure}
\hfill
\begin{subfigure}{0.49\textwidth}
  \centering
  \includegraphics[width=1\linewidth]{pics/3dshapes_factorvae_rtd_nontrivial.png}
    \caption{FactorVAE + TopDis, \mbox{3D Shapes}.}
    \label{fig:FactorVAE_RTD_3dshapes_traverse}
\end{subfigure}

\begin{subfigure}{0.49\textwidth}
  \centering
  \includegraphics[width=1\linewidth]{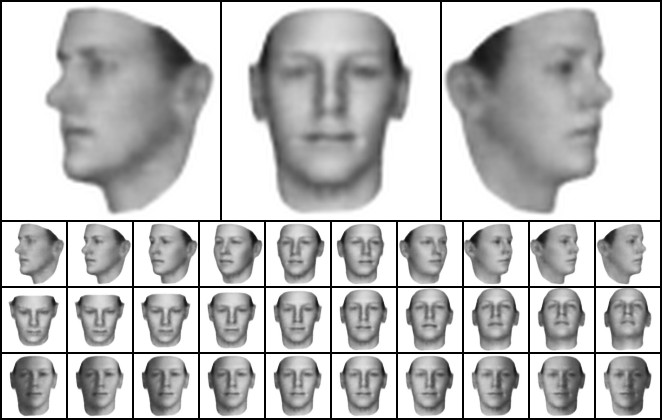}
    \caption{FactorVAE, \mbox{3D Faces}.}
    \label{fig:FactorVAE_3dfaces_traverse}
\end{subfigure}
\hfill
\begin{subfigure}{0.49\textwidth}
  \centering
  \includegraphics[width=1\linewidth]{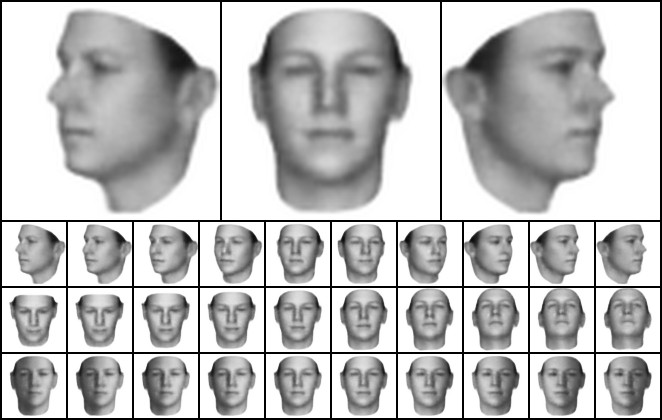}
    \caption{FactorVAE + TopDis, 3D Faces.}
    \label{fig:FactorVAE_RTD_3dfaces_traverse}
\end{subfigure}

\begin{subfigure}{0.49\textwidth}
  \centering
  \includegraphics[width=1\textwidth]{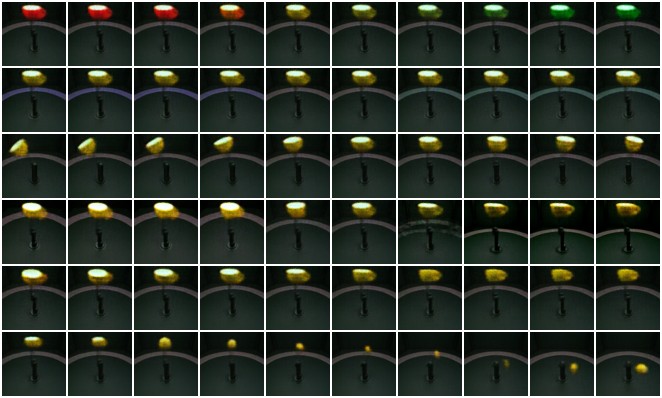}
    \caption{FactorVAE, \mbox{MPI 3D}.}
    \label{fig:FactorVAE_MPI3D_traverse}
\end{subfigure}
\hfill
\begin{subfigure}{0.49\textwidth}
  \centering
  \includegraphics[width=1\textwidth]{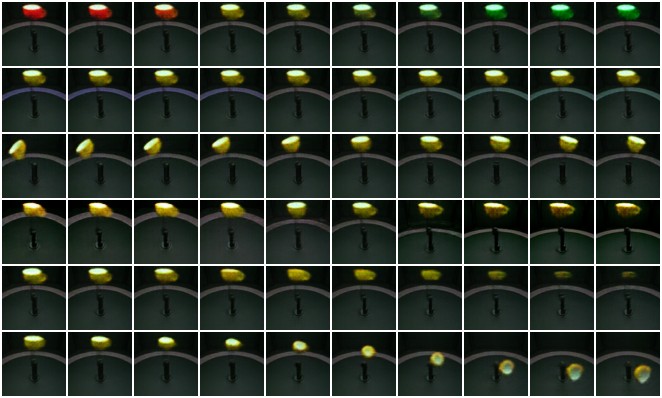}
    \caption{FactorVAE + TopDis, MPI 3D.}
    \label{fig:FactorVAE_RTD_MPI3D_traverse}
\end{subfigure}

\caption{FactorVAE and FactorVAE + TopDis latent traversals.}
\label{fig:latent_traversals}
\end{figure}

\begin{figure}[!ht]
\centering
\begin{subfigure}{0.49\textwidth}
  \centering
  \includegraphics[width=1\linewidth]{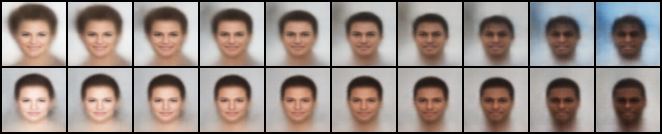}
    \caption{Skin tone}
    \label{fig:celeba_skin}
\end{subfigure}
\hfill
\begin{subfigure}{0.49\textwidth}
  \centering
  \includegraphics[width=1\linewidth]{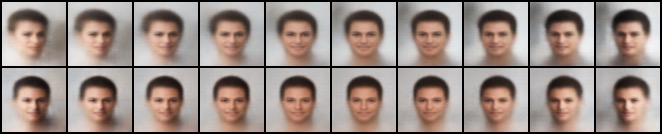}
    \caption{Lightning}
    \label{fig:celeba_light}
\end{subfigure}
\caption{Visual improvement from addition of TopDis, CelebA. Top: FactorVAE, bottom: FactorVAE + TopDis.}
\label{fig:celeba_main}
\end{figure}

Images obtained from selected latent traversal exhibiting the most differences are presented in Figure \ref{fig:latent_traversals} (FactorVAE, FactorVAE+TopDis trained on dSprites, 3D shapes, MPI 3D, 3D Faces) and Figure \ref{fig:celeba_main} (FactorVAE, FactorVAE+TopDis trained on CelebA).

\begin{figure}
\centering
\begin{subfigure}{0.49\textwidth}
  \centering
  \includegraphics[width=0.99\linewidth]{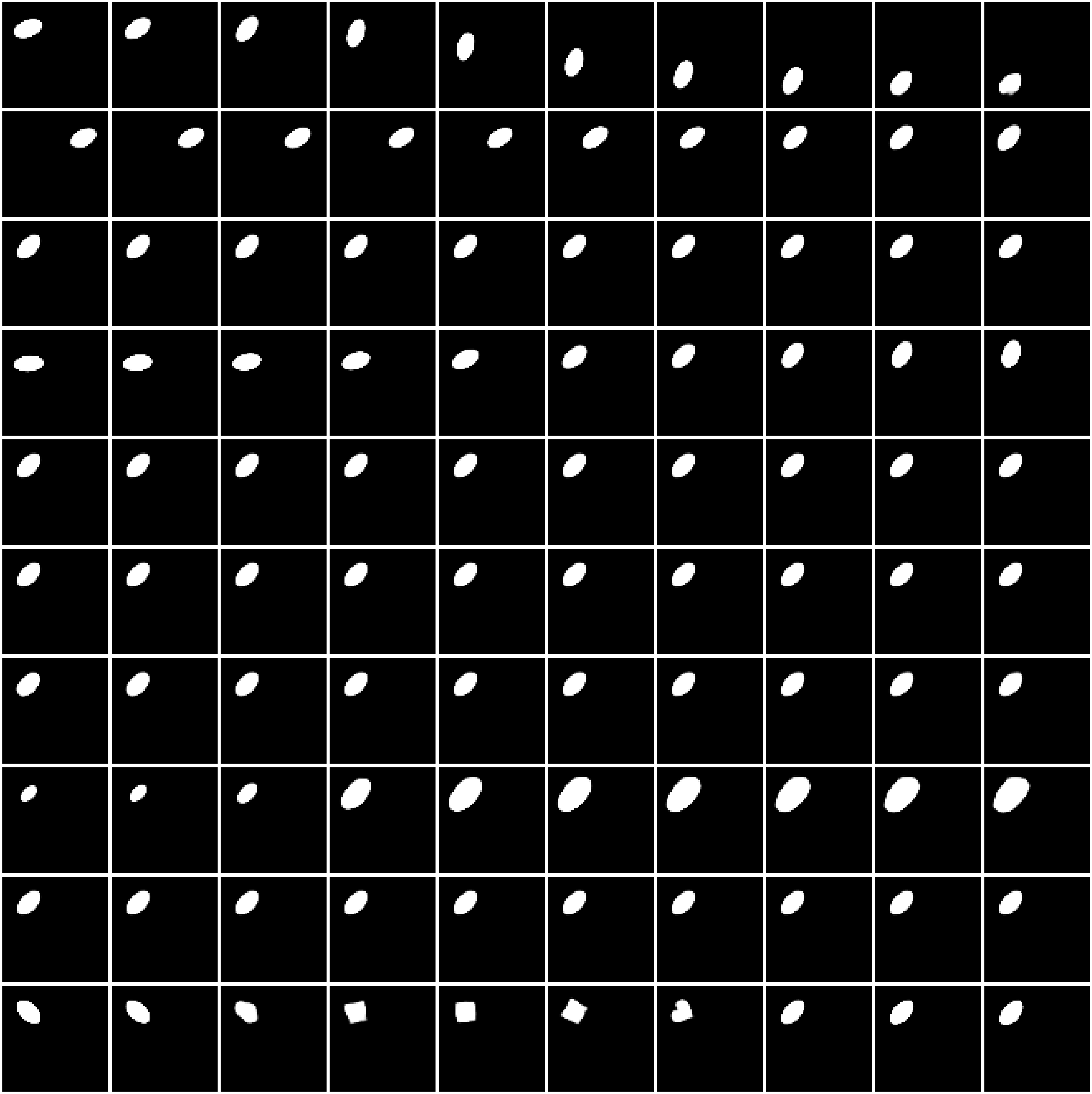}
\end{subfigure}
\hfill
\begin{subfigure}{0.49\textwidth}
  \centering
  \includegraphics[width=0.99\linewidth]{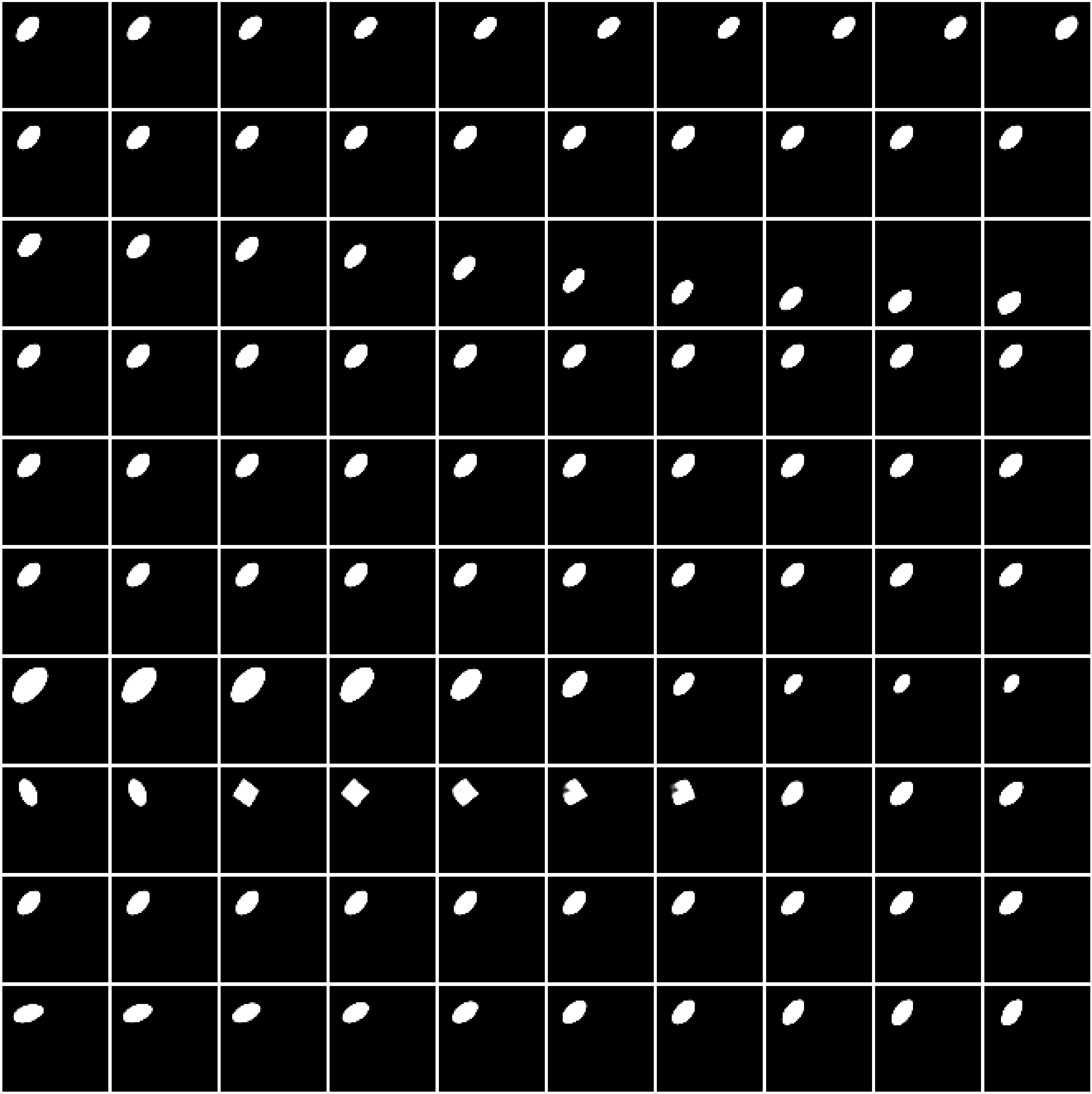}
\end{subfigure}
\caption{FactorVAE (left) and FactorVAE + TopDis (right) latent traversals, \mbox{dSprites}.}
\label{fig:latent_traversals_full_dsprites}
\end{figure}

\begin{figure}
\centering
\begin{subfigure}{0.49\textwidth}
  \centering
  \includegraphics[width=0.99\linewidth]{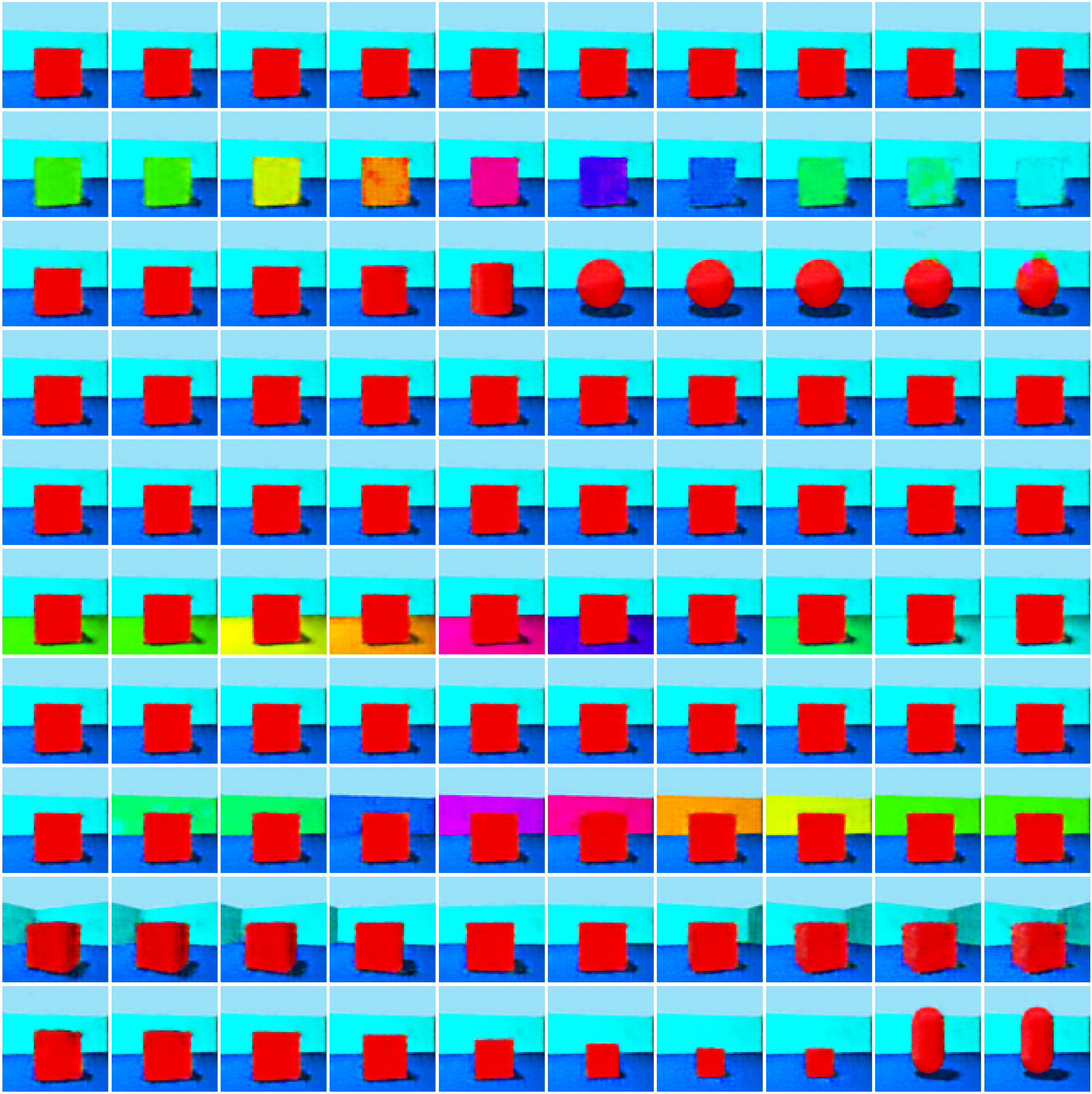}
\end{subfigure}
\hfill
\begin{subfigure}{0.49\textwidth}
  \centering
  \includegraphics[width=0.99\linewidth]{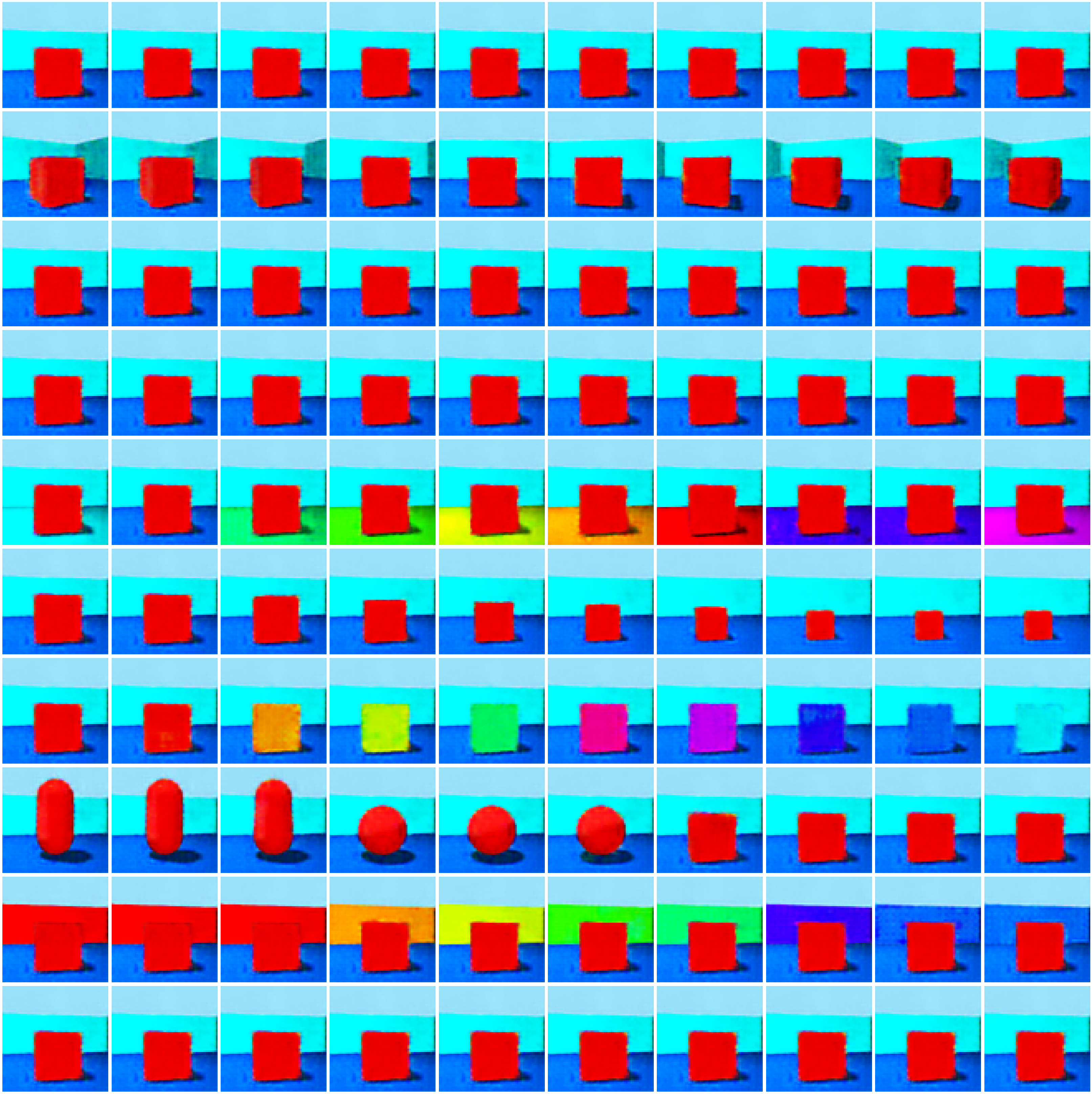}
\end{subfigure}
\caption{FactorVAE (left) and FactorVAE + TopDis (right) latent traversals, \mbox{3D Shapes}.}
\label{fig:latent_traversals_full_3dshapes}
\end{figure}

\begin{figure}
\centering
\begin{subfigure}{0.49\textwidth}
  \centering
    \includegraphics[width=0.99\textwidth]{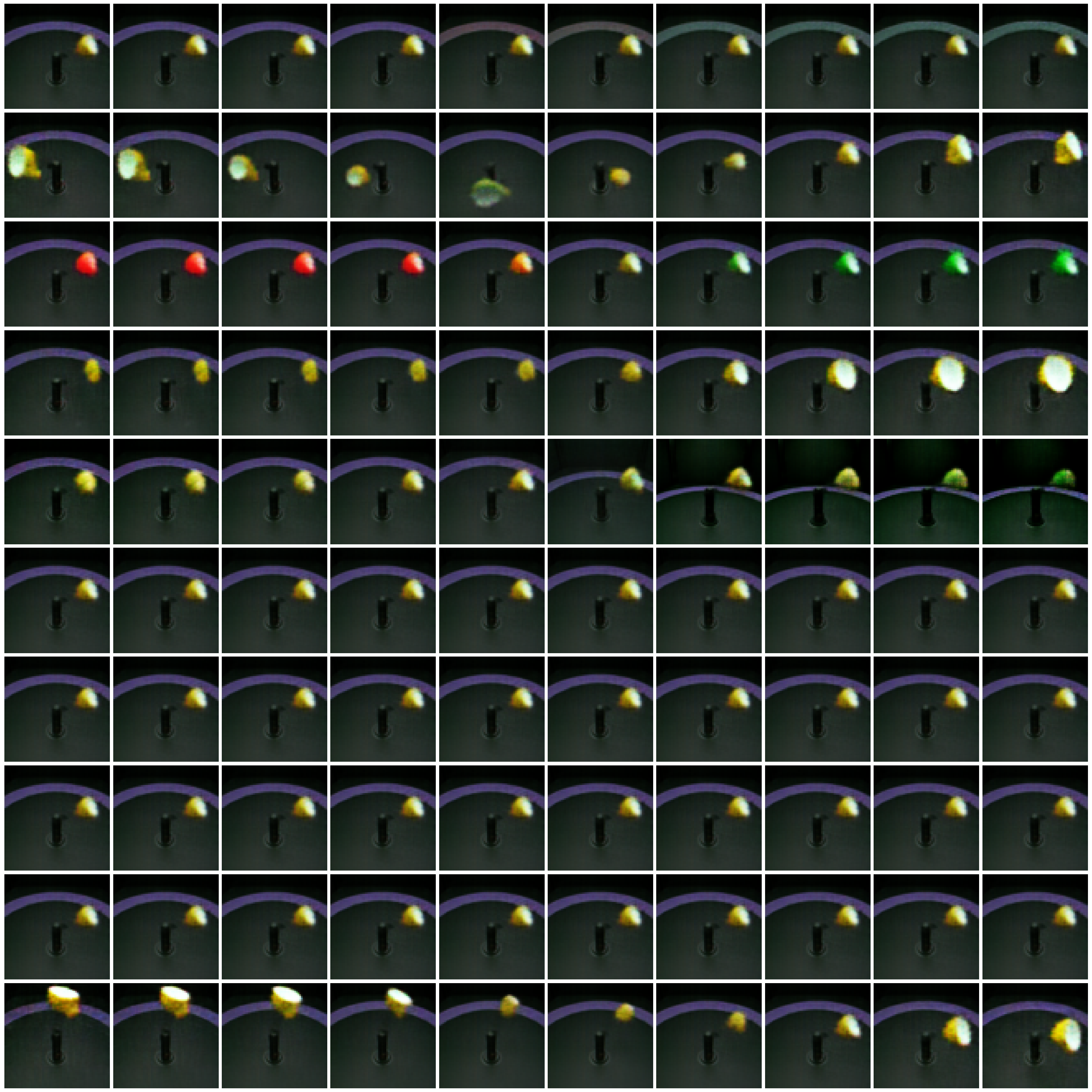}
\end{subfigure}
\hfill
\begin{subfigure}{0.49\textwidth}
  \centering
    \includegraphics[width=0.99\textwidth]{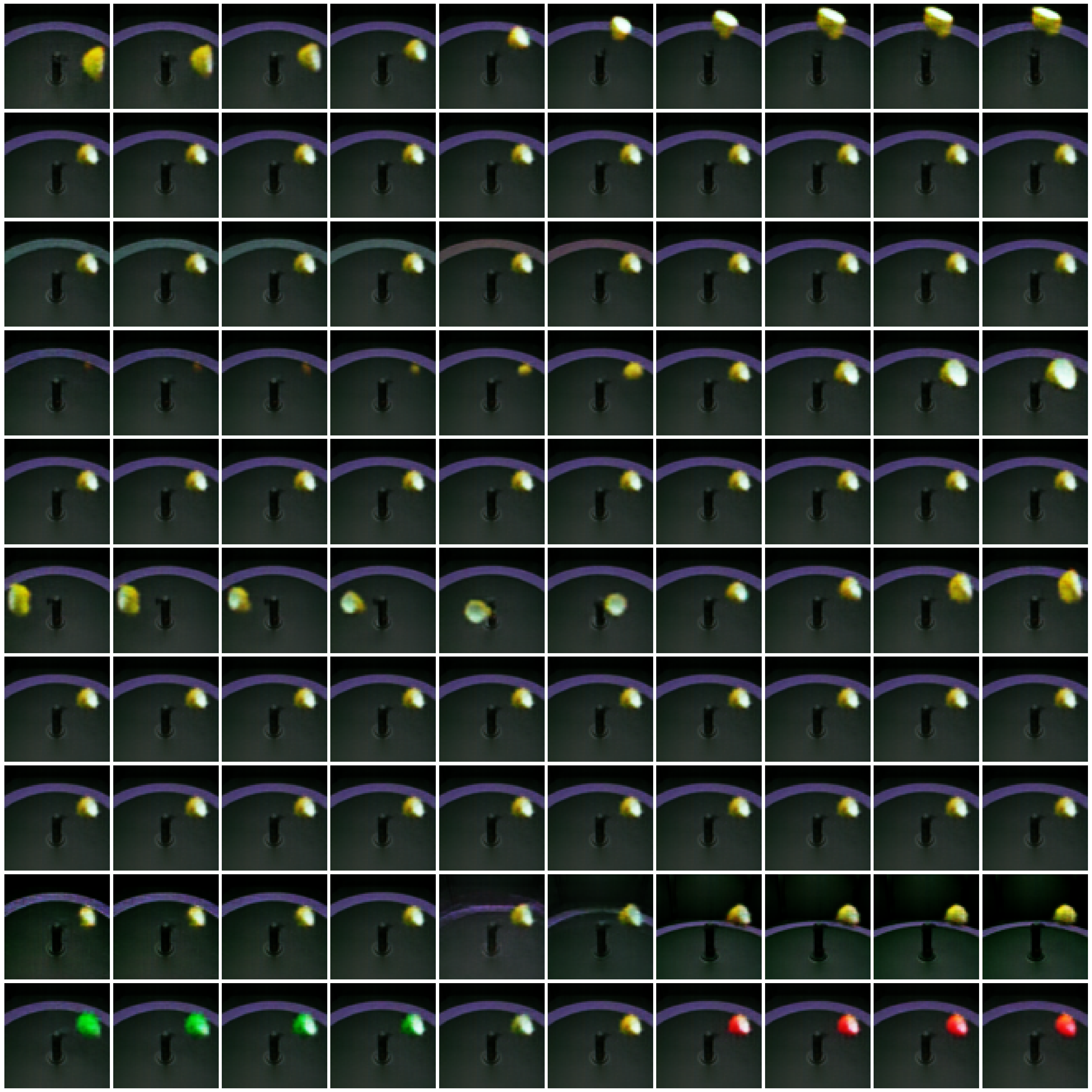}
\end{subfigure}
\caption{FactorVAE (left) and FactorVAE + TopDis (right) latent traversals, \mbox{MPI 3D}.}
\label{fig:latent_traversals_full_mpi3d}
\end{figure}

\begin{figure}
\centering
\begin{subfigure}{0.49\textwidth}
  \centering
  \includegraphics[width=0.99\linewidth]{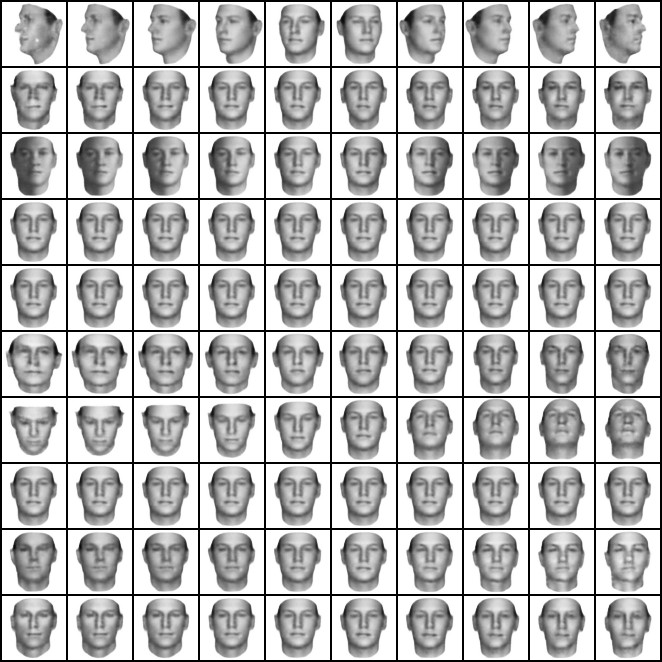}
\end{subfigure}%
\hfill
\begin{subfigure}{0.49\textwidth}
   \centering
  \includegraphics[width=0.99\linewidth]{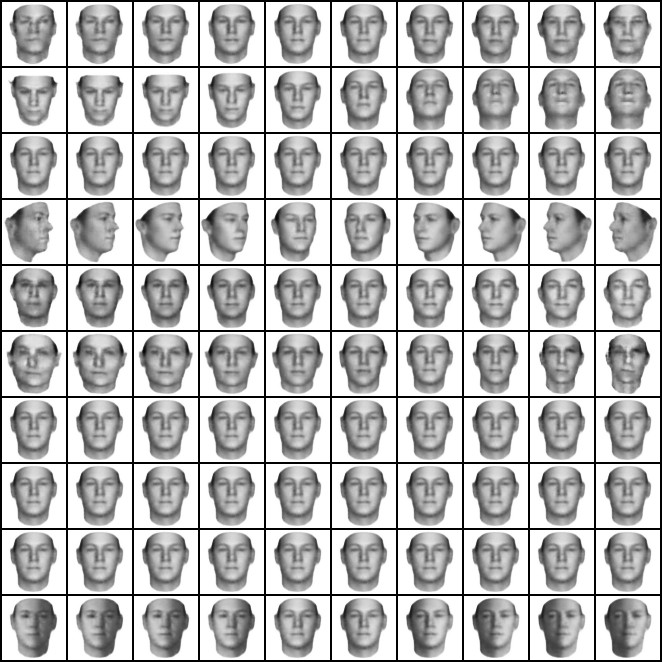}
\end{subfigure}
\caption{FactorVAE (left) and FactorVAE + TopDis (right) latent traversals, \mbox{3D Faces}.}
\label{fig:latent_traversals_full_2}
\end{figure}

\begin{figure}
\centering
\begin{subfigure}{0.49\textwidth}
  \centering
  \includegraphics[width=0.99\linewidth]{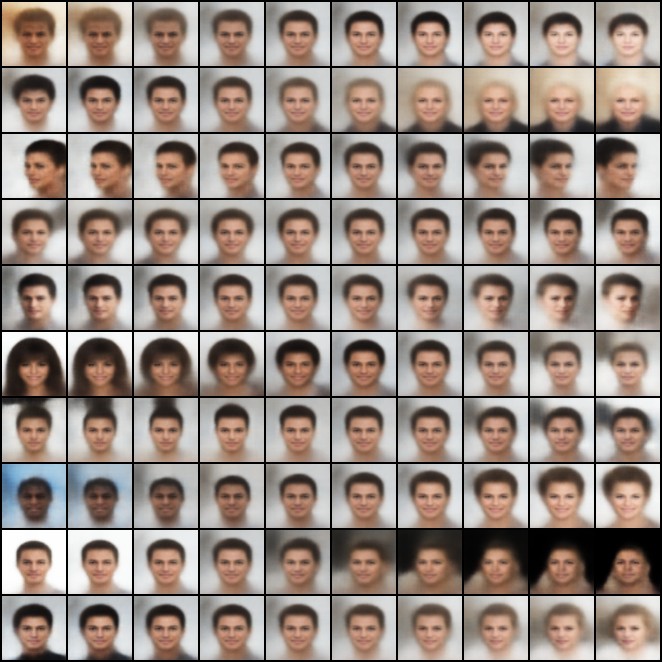}
\end{subfigure}
\hfill
\begin{subfigure}{0.49\textwidth}
  \centering
  \includegraphics[width=0.99\linewidth]{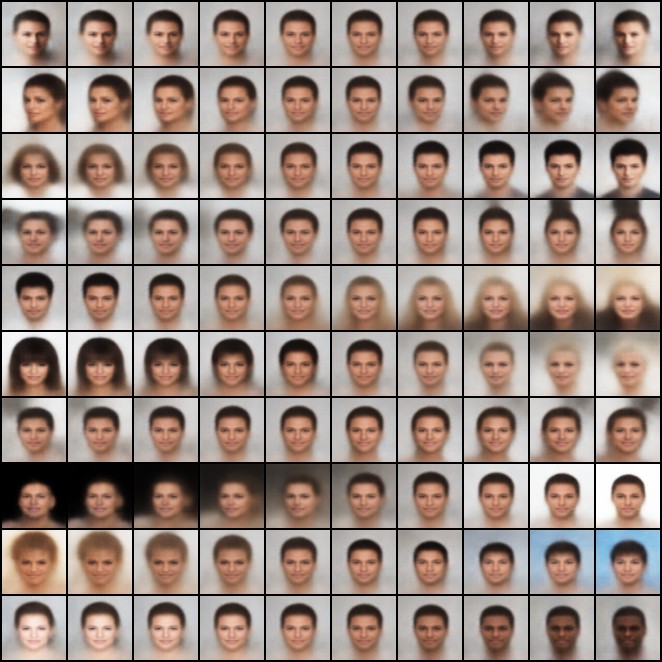}
\end{subfigure}
\caption{FactorVAE (left) and FactorVAE + TopDis (right) latent traversals, CelebA.}
\label{fig:latent_traversals_full_3}
\end{figure}

Figures \ref{fig:latent_traversals_full_dsprites}, \ref{fig:latent_traversals_full_3dshapes}, \ref{fig:latent_traversals_full_mpi3d}, \ref{fig:latent_traversals_full_2}, \ref{fig:latent_traversals_full_3} shows latent traversals along all axes.

\textbf{dSprites.} 
Figures \ref{fig:FactorVAE_dsprites_traverse} and \ref{fig:FactorVAE_RTD_dsprites_traverse} show that the TopDis loss helps to outperform simple FactorVAE in terms of visual perception. The simple FactorVAE model has entangled rotation and shift along axes (rows 1,2,5 in Figure \ref{fig:FactorVAE_dsprites_traverse}), even though the Total Correlation in both models is minimal, which demonstrates the impact of the proposed topological objective.

\textbf{3D Shapes.} 
Figures 
\ref{fig:FactorVAE_3dshapes_traverse} and \ref{fig:FactorVAE_RTD_3dshapes_traverse}
show that proposed TopDis loss leads to the superior disentanglement of the factors as compared to the simple FactorVAE model, where the shape and the scale factors remain entangled in the last row.

\textbf{3D Faces.} FactorVAE+TopDis (Figure \ref{fig:FactorVAE_RTD_3dfaces_traverse}) outperforms FactorVAE (Figure \ref{fig:FactorVAE_3dfaces_traverse}) in terms of disentangling the main factors such as azimuth, elevation, and lighting from facial identity. On top of these figures we highlight the azimuth traversal. The advantage of TopDis is seen from the observed preservations in facial attributes such as the chin, nose, and eyes.

\textbf{MPI 3D.} Here, the entanglement between the size and elevation factors is particularly evident when comparing the bottom two rows of Figures \ref{fig:FactorVAE_MPI3D_traverse} and \ref{fig:FactorVAE_RTD_MPI3D_traverse}. In contrast to the base FactorVAE, which left these factors entangled, our TopDis method successfully disentangles them.

\textbf{CelebA.} For this dataset, we show the most significant improvements obtained by adding the TopDis loss in Figure \ref{fig:celeba_main}. The TopDis loss improves disentanglement of skin tone and lightning compared to basic FactorVAE, where these factor are entangled with other factors - background and hairstyle.

%
%

\end{document}